\documentclass{article} % For LaTeX2e
\usepackage{iclr2026_conference,times}

\usepackage{amsmath,amsfonts,bm}

\def\eqref#1{equation~\ref{#1}}
\def\1{\bm{1}}

\DeclareMathAlphabet{\mathsfit}{\encodingdefault}{\sfdefault}{m}{sl}
\SetMathAlphabet{\mathsfit}{bold}{\encodingdefault}{\sfdefault}{bx}{n}

\usepackage{hyperref}
\usepackage{url}
\usepackage{amsfonts}
\usepackage{amssymb}
\usepackage{multirow}
\usepackage{booktabs}
\usepackage{graphicx}
\usepackage{algorithm}
\usepackage{algpseudocode}
\usepackage{dsfont}
\usepackage{subfig}
\usepackage{adjustbox}
\usepackage{multicol}
\usepackage{subcaption}
\usepackage{xcolor}
\title{Enhancing Trust in Large Language Models via Uncertainty-Calibrated Fine-tuning}
\iclrfinalcopy
\author{Ranganath Krishnan \thanks{Work done while at Intel.} \\
Capital One, AI Labs\\
Texas, USA \\
\texttt{\tiny{ranganath.krishnan@capitalone.com}}
\And
Piyush Khanna \footnotemark[1] \\
Wayve Technologies \\
California, USA \\
\texttt{\tiny{piyush.khanna@wayve.ai}} \\
\And
Omesh Tickoo \\
Intel Corporation \\
Oregon, USA \\
\texttt{\tiny{omesh.tickoo@intel.com}}
}

\begin{document}

\maketitle

\begin{abstract}
Large language models (LLMs) have achieved remarkable success in natural language generation, yet they remain prone to hallucinations and often exhibit miscalibrated overconfidence, even when producing incorrect outputs. Reliable uncertainty estimation is a key requirement for deploying LLMs safely, as it enables users and downstream systems to assess confidence, detect hallucinations, and identify out-of-domain prompts. In this work, we propose an uncertainty-calibrated fine-tuning approach that improves the reliability of LLMs in open-ended, free-form generation settings. Our method introduces a novel uncertainty-aware causal language modeling loss, grounded in decision-theoretic principles, that explicitly encourages well-calibrated uncertainty estimates. We conduct extensive empirical evaluations across multiple free-form question-answering datasets and model architectures, demonstrating that our approach consistently yields better uncertainty calibration compared to standard fine-tuning.
Furthermore, the experimental results show that
the proposed method substantially enhances the model’s ability to detect hallucinations and identify out-of-domain prompts.

\end{abstract}

\section{Introduction}
\label{sec:introduction}
Large Language Models (LLMs) have shown remarkable success in various natural language processing tasks \citep{llama2023touvron, gemma2024, achiam2023gpt} and are increasingly becoming ubiquitous in a variety of domains for their decision-making and reasoning abilities \citep{eigner2024determinants}. However, their real-world deployment, particularly in high-stakes and safety-critical applications, is hindered by challenges such as hallucinations and out-of-domain prompts, which can lead to the generation of erroneous or nonsensical outputs. LLM Hallucinations, which are plausible-sounding but incorrect model generations \citep{kalai2025language, ji2023survey}, present a crucial challenge in developing trustworthy AI systems. The ability to recognize out-of-domain prompts and to acknowledge the limits of a model’s knowledge base paves the way for building safe AI systems \citep{amodei2016concrete, bhatt2021uncertainty}. 

Uncertainty quantification (UQ) in LLMs plays a pivotal role in understanding what the model knows and does not know, which is an active area of research for free-form natural language generation (NLG)~\citep{kadavath2022language, kuhn2023semantic, lin2024generating}. UQ methods has emerged as a step towards 
determining the trustworthiness of responses generated by LLMs~\citep{fadeeva2023lm, plaut2024softmax, kadavath2022language}. Uncertainty estimation techniques such as semantic entropy \citep{kuhn2023semantic} have shown to be effective indicators in detecting `confabulations'~\citep{farquhar2024detecting}, a subcategory of hallucinations characterized by the generation of arbitrary and incorrect responses.

Calibration of uncertainty estimates is crucial for the reliability of LLMs; a well-calibrated model should correlate low uncertainty with accurate responses and high uncertainty with incorrect responses. However, recent studies \citep{xiong2024can, yang2024can, groot2024overconfidence} have revealed that LLM predictions are often poorly calibrated, leading to overconfidence in incorrect outputs. This problem is more pronounced in fine-tuned language models~\citep{kong2020calibrated, litcab2024}. Unlike the pre-training phase, where models are exposed to vast amounts of unlabeled data, fine-tuning involves limited labeled data. Consequently, the immense capacity of LLMs can lead to overfitting on this limited data, producing overconfident predictions \citep{kong2020calibrated}. This presents a substantial challenge, as the model may produce unreliable uncertainty metrics that are influenced by the model's miscalibrated token probability. 
%Moreover, evaluating calibration in NLG is particularly challenging due to the variable lengths of generated text compared to reference sentences. Traditional calibration error metrics~\citep{naeini2015obtaining, nixon2019measuring} assume a fixed number of outcomes or classes, which aligns well with classification tasks but not with the open-ended nature of NLG, where the number of tokens in generated text and reference sentences can differ. To address this, we exploit the inverse correlation between uncertainty quantification and the quality of generated text, allowing us to perform calibration analysis in NLG settings.

\citet{kalai2025language} argue that hallucinations in large language models are an inevitable consequence of statistical training objectives and evaluation practices that reward guessing over expressing uncertainty, and that addressing this failure mode requires fundamentally rethinking misaligned benchmarks to prioritize uncertainty estimation. While their work centers on reforming evaluation practices, our work directly advances this agenda by proposing a method that enable LLMs to be trustworthy with well-calibrated uncertainty estimates, which can be leveraged to more effectively detect hallucinations.

In this paper, we present a novel method for fine-tuning LLMs within a causal language modeling framework that incorporates uncertainty calibration, based on principles from decision theory. Our approach is orthogonal to existing uncertainty quantification methods and is driven by the goal of enhancing the reliability of those UQ metrics through uncertainty-calibrated fine-tuning. Specifically, we achieve this with an optimization objective that encourages the model to learn to associate high uncertainty with incorrectly generated tokens and low uncertainty with correctly generated tokens, while maximizing accuracy within the framework of causal language modeling. We show that fine-tuning with our calibration objective enhances the reliability of uncertainty quantification in LLMs. Calibrated uncertainty estimates serve as a crucial tool for enhancing the trustworthiness of generated responses, represents a substantial step towards identifying hallucinations, and improving decision-making capabilities through selective generation~\citep{ren2022out}. 

Our main contributions in this paper are as follows:
\begin{itemize}
\item We introduce a novel uncertainty-aware causal language modeling (UA-CLM) loss function, grounded in decision theory, designed to produce well-calibrated uncertainty estimates in free-form natural language generation.
%\item We conduct sentence-level and token-level uncertainty calibration analysis in free-form NLG settings, employing an innovative methodology that leverages the inverse correlation between uncertainty quantification and the quality of the generated text. 
\item We conduct a comprehensive empirical evaluation across four key aspects: hallucination detection, selective generation, out-of-prompt detection, and calibration analysis. Our findings demonstrate that UA-CLM significantly improves the quality of uncertainty estimates in LLMs for free-form text generation tasks.
%Notably, these enhancements in uncertainty calibration are achieved without compromising the accuracy when compared to standard CLM.
%\item In addition, we have also applied our proposed UA-CLM methodology to a large vision-language model (LVLM), demonstrating its efficacy in the open-ended visual question-answering task. This extension of our work shows the versatility of UA-CLM in handling multimodal inputs and complex tasks beyond text-based question-answering, showcasing it's utilty to wider range of applications.
%\item We investigate sentence-level uncertainty calibration analysis in free-form NLG settings using an innovative method leveraging the inverse correlation between uncertainty quantification and text quality.
%\item We applied our UA-CLM methodology to a large vision-language model, demonstrating its effectiveness in handling multimodal inputs and complex tasks beyond text-based question-answering.
\item We demonstrate the effectiveness of UA-CLM methodology on a vision-language model, highlighting its ability to generalize to multimodal models and complex tasks.
\end{itemize}

\section{Background and Related works}
\label{sec:background}

\subsection{Uncertainty estimation in LLMs}

For comprehensive surveys on uncertainty quantification (UQ) in deep neural networks and LLMs, we refer to \cite{abdar2021review, gawlikowski2023survey, liu2025uncertainty}. As LLMs evolve, understanding the uncertainty in their responses is crucial for developing trustworthy systems \citep{fadeeva2023lm}. While UQ methods from deep learning are effective for tasks like text classification \citep{xiao2019quantifying} and multiple-choice question answering \citep{kumar2023conformal}, applying them to free-form NLG and open-ended tasks presents unique challenges.

The landscape of UQ in free-form NLG is a dynamic field of research, generally bifurcated into white-box \citep{fomicheva2020unsupervised, kuhn2023semantic, kossen2024semantic, shelmanov2025head, ni2026reasoning} and black-box approaches \citep{lin2024generating, xiong2024can, zhang-etal-2024-luq}. White-box methods require access to model internals, while black-box methods analyze generated text sequences. Prompting techniques \citep{tian2023just} have been explored to elicit verbalized response confidence in LLMs. Conformal prediction methods have been proposed for uncertainty quantification in large language models \citep{vishwakarmaprune, tayebati2025learning}. Unsupervised methods use confidence \citep{plaut2024softmax}, perplexity \citep{fomicheva2020unsupervised}, token entropy \citep{malininuncertainty}, semantic entropy \citep{kuhn2023semantic}, and pretrained UQ heads \citep{shelmanov2025head} to quantify uncertainty in LLM responses. Prior studies have systematically benchmarked the performance of different UQ methods in large language models \citep{vashurin2024benchmarking, bouchard2025uncertainty}. However, prior works \citep{minderer2021revisiting, xiong2024can} indicate that confidence or entropy measures can suffer from poor calibration. Our work aims to improve these metrics through uncertainty-aware fine-tuning, enhancing the calibration of language models.

%The landscape of UQ in free-form NLG is a dynamic field of research, where methodologies are generally bifurcated into white-box~\citep{fomicheva2020unsupervised,kuhn2023semantic} and black-box approaches~\citep{lin2024generating, xiong2024can}. White-box methods necessitate access to the model's logits, or likelihood scores, or other internals of LLMs. Black-box methods rely solely on the analysis of the text sequences generated by the LLMs.  Prompting techniques \citep{tian2023just} have been explored to explicitly elicit verbalized response confidence in LLMs. Another line of prominent research involves unsupervised methods for UQ from the models by utilizing confidence \citep{plaut2024softmax}, perplexity \citep{fomicheva2020unsupervised}, token entropy~\citep{malininuncertainty} and semantic entropy \citep{kuhn2023semantic} to quantify uncertainty in LLM responses. 
%Prior works \citep{minderer2021revisiting, xiong2024can} have shown that confidence or entropy measures can be susceptible to poor calibration and may not fully reflect a model's underlying uncertainties. Our work focuses on improving these uncertainty metrics in white-box settings by better calibrating the language models with uncertainty-aware fine-tuning.

\subsection{Model Calibration}
Calibration is important in applications where decision-making relies not just on the predictions of the model, but also on the trustworthiness of its uncertainty scores. It is important to capture well-calibrated uncertainty estimates to create trustworthy systems. In deep learning, model calibration has been extensively researched, with strategies like post-hoc rescaling~\citep{guo2017calibration, kull2017beta}, data augmentation~\citep{thulasidasan2019mixup, hendrycksaugmix}, and probabilistic modeling \citep{blundell2015weight, lakshminarayanan2017simple, krishnan2020specifying} enhancing calibration for classification and regression tasks. Explicit calibration loss functions~\citep{kumar2018trainable, krishnan2020improving, karandikar2021soft} have also improved model calibration.

For LLMs, calibration remains an active research area \citep{geng2024survey, tao2025revisiting, liu2025uncertainty}. Despite their impressive performance, foundational LLMs often show poor calibration, particularly overconfidence \citep{xiong2024can, kong2020calibrated, desai2020calibration, jiang2021can}. Techniques such as Unlikelihood Training (ULT) \citep{welleckneural} and Calibration Tuning (CT) \citep{kapoor2024calibration} have been proposed to address these issues. Other methods include adjusting model's logits \citep{litcab2024}, learning auxiliary models \citep{shenthermometer}, instruction tuning \cite{liu-etal-2024-llms-learn-uncertainty}, and using reinforcement learning \citep{band2024linguistic}. Standard fine-tuning of LLMs can lead to poorer calibration \citep{litcab2024}, highlighting the need for improved methods. Calibration of LLMs for free-form text generation, as well as uncertainty-aware fine-tuning, represents a significant open area of research. Our work contributes to this field by developing an uncertainty-aware causal language modeling method via parameter-efficient fine-tuning of LLMs, aiming for well-calibrated uncertainty quantification in free-form text generation. By focusing on causal language modeling, our approach addresses calibration issues and enhances the generalization of uncertainty calibration in LLMs.

\subsection{Fine-tuning Large Language Models}
With the emergence of foundation models, fine-tuning has become essential for adapting general-purpose pre-trained models to specialized tasks and domains. Given the resource-intensive nature of full fine-tuning for LLMs with billions of parameters, parameter-efficient fine-tuning (PEFT) strategies have emerged \citep{peft, han2024parameter}. These techniques mitigate catastrophic forgetting more effectively than full fine-tuning \citep{wang2022adamix}. Recent studies \citep{gekhman2024does} indicate that full fine-tuning can exacerbate LLM hallucinations, making PEFT strategies crucial. PEFT involves updating only a subset of model parameters, such as adapter modules \citep{houlsby2019parameter} or Low-Rank Adaptation (LoRA) \citep{lora2022}, while keeping pre-trained weights frozen. Prompt-based fine-tuning \citep{liu2023pre} conditions models on task-specific prompt embeddings without updating model parameters. In this work, we utilize LoRA to implement our uncertainty-aware causal language modeling, fine-tuning less than 1\% of model parameters. This approach helps preserve the pre-trained model's knowledge, reducing overfitting and maintaining the integrity of the foundation model.
%With the emergence of foundation models, fine-tuning have become a common practice to enable the adaptation of general-purpose pre-trained models to specialized task and domains. As full fine-tuning of LLMs with billions of parameters can be resource-intensive, parameter-efficient fine-tuning~\citep{peft, han2024parameter} strategies have evolved. These parameter-efficient fine-tuning (PEFT) techniques also mitigate catastrophic forgetting more effectively in comparison to full fine-tuning \citep{wang2022adamix}. Recent work by \citet{gekhman2024does} has shown that full fine-tuning of LLMs, where all model parameters are updated, can exacerbate LLM hallucinations. PEFT strategies can help to overcome these challenges. PEFT approaches involve updating only a subset of the model's parameters, such as adapter modules~\citep{houlsby2019parameter} or Low-Rank Adaptation (LoRA)~\citep{lora2022}, where only a small set of parameters are updated while the pre-trained weights are frozen. Another strategy is prompt-based fine-tuning \citep{liu2023pre}, where models are conditioned on task-specific prompt parameters to learn without model parameter updates. In this work, we leverage LoRA to illustrate our proposed uncertainty-aware causal language modeling, fine-tuning less than 1\% of the model parameters. This approach helps preserve the pre-trained model's knowledge and reduces the risk of overfitting and conflicts, thereby maintaining the integrity of the foundation model.

\section{Uncertainty-aware Causal Language Modeling}
\label{sec:proposed_method}

Motivated by the need to overcome the challenges of uncertainty miscalibration~\citep{xiong2024can} in Large Language Models (LLMs) and the increasing trend of fine-tuning pre-trained foundational models for domain-specific adaptation — where fine-tuned LLMs often exhibit overconfidence in their predictions~\citep{kong2020calibrated} $\hbox{—}$ we propose a novel uncertainty calibration fine-tuning approach for natural language generation settings. We introduce a novel uncertainty-aware causal language modeling loss based on the principles of decision theory~\citep{murphy2012machine}. Our fine-tuning approach emphasizes increasing the uncertainty for wrong token predictions, while optimizing for accuracy and certainty for correct token predictions. Decision theory offers a mathematical and theoretical framework that guides to achieve optimal predictions by employing a task-specific utility function. Within the decision theory framework, our task is to generate natural language text accompanied by reliable uncertainty estimates. The utility function in this scenario is represented by the uncertainty-aware optimization objective function that is aimed at producing well-calibrated uncertainty estimates for causal language modeling. We design a differentiable loss function that incentivizes the model to yield low uncertainty when it generates correct tokens, and encourages the model to exhibit high uncertainty when it is at risk of predicting the next token incorrectly.

In causal language modeling, the goal is to predict the next token in a sequence given the previous tokens. Given a sequence of tokens $[w_1, w_2, \ldots, w_T ]$, where T is the length of the sequence and each token $w_i$ is an element from a fixed vocabulary of size V, the model aims to learn the conditional probability distribution $P_{\theta}(w_i | w_{0:i\hbox{-}1})$ for each token $w_i$ given the preceding set of tokens $w_{0:i\hbox{-}1}$; where, $\theta$ represents the parameters of the LLM. The loss function for standard causal language modeling (CLM) is typically the negative log-likelihood as defined in Equation~\ref{eqn:clm} below.
\begin{equation}
\small
\label{eqn:clm}
\begin{aligned}
\mathcal{L^{\theta}_{\mathrm{CLM}}} &:= -\frac{1}{T} \sum_{i=0}^{T}  \log P_{\theta}(w_{i} | w_{0:i-1})
\end{aligned}
\end{equation}
\paragraph{Desideratum:} The desired and ideal outcome in causal language modeling is to achieve a state where every correctly generated token is assigned low predictive uncertainty, and high predictive probability, reflecting the model's high confidence in its accuracy. Conversely, for every token that is generated incorrectly, the model should assign high uncertainty, and low predictive probability, denoting low confidence in these instances. This ensures that the model's confidence levels and uncertainty estimates are perfectly calibrated with the actual correctness of its predictions.

We define the uncertainty-aware causal language modeling (UA-CLM) loss in Equation~\ref{eqn:uaclm_loss} based on the above desideratum. The loss function captures the trade-off between predictive accuracy and uncertainty calibration, and is composed of two utilities: one that deals with incorrectly generated tokens and one that deals with correctly generated tokens.
\begin{equation}
\small
\label{eqn:uaclm_loss}
\begin{aligned}
%\mathcal{L^{\theta}_{\mathrm{UA\hbox{-}CLM}}} &:= \underbrace{-\frac{1}{\sum_i \mathds{1}\left(w_i \neq \bar{w}_i\right)}\sum_{i \in \{w_i \neq \bar{w_i}\}} P_{\theta}(w_i | w_{0: i\hbox{-}1}) \log \left(\tanh \left(H_i\right)\right)}_\textrm{Utility function for incorrect tokens}  \\
%& \qquad \qquad \underbrace{-\frac{1}{\sum_i \mathds{1}\left(w_i=\bar{w}_i\right)} \sum_{i \in \{w_i=\bar{w}_i\}}\left(1-P_{\theta}(w_i | w_{0: i\hbox{-}1})\right) \log \left(1-\tanh \left(H_i\right)\right)}_\textrm{Utility function for correct tokens}
\mathcal{L^{\theta}_{\mathrm{UA\hbox{-}CLM}}} &:= \underbrace{-\frac{1}{|\widetilde{C}|}\sum_{i \, \in \, \widetilde{C}} P_{\theta}(w_i | w_{0: i\hbox{-}1}) \log \left(\tanh \left(H_i\right)\right)}_\textrm{Utility function for incorrect tokens}  \\ 
&\underbrace{-\frac{1}{|C|} \sum_{i \, \in \, C}\left(1-P_{\theta}(w_i | w_{0: i\hbox{-}1})\right) \log \left(1-\tanh \left(H_i\right)\right)}_\textrm{Utility function for correct tokens}
\end{aligned}
\end{equation}
%\begin{equation}
%\label{eqn:uaclm_loss_singleline}
%\mathcal{L^{\theta}_{\mathrm{UA\hbox{-}CLM}}} &:= \underbrace{-\frac{1}{\widetilde{C}}\sum_{i \in \widetilde{C}} P_{\theta}(w_i | w_{0: i\hbox{-}1}) \log \left(\tanh \left(H_i\right)\right)}_\textrm{Utility function for incorrect tokens}   \underbrace{-\frac{1}{C} \sum_{i \in C}\left(1-P_{\theta}(w_i | w_{0: i\hbox{-}1})\right) \log \left(1-\tanh \left(H_i\right)\right)}_\textrm{Utility function for correct tokens}
%\end{equation}
where,
 \begin{equation}
 \small
H_i:= -\sum_{j=1}^{V} P_{\theta}(w_{i}^{j} | w_{0: i\hbox{-}1}) \log P_{\theta}(w_{i}^{j} | w_{0: i\hbox{-}1})
\end{equation}
%$argmax\left(P_{\theta}(w_i | w_{0: i\hbox{-}1})\right)$

Here, $C := \{ i \mid w_i=\overline{{w}_i}\}$ is a set of indices corresponding to correctly generated tokens, $\widetilde{C} := \{ i \mid w_i \neq \overline{w_i}\}$ is a set of indices corresponding to incorrectly generated tokens in the mini-batch during fine-tuning, ${\overline{w_i}}$ is the ground-truth reference token, $w_i$ is the $argmax$ of the of the predictive probability distribution, $H_i$ is the token entropy of the probability distribution of the $i^{th}$ token in the sequence, $P_{\theta}(w_{i}^{j} | w_{0: i\hbox{-}1})$ is the predicted probability of the $j^{th}$ token in the vocabulary of size V. The hyperbolic tangent function is employed to scale the token entropy values, ensuring that  $\tanh(H_i$) lies in the interval [0, 1]. We used the hyperbolic tangent function for its smooth gradient properties following prior works in uncertainty calibration \citep{krishnan2020improving, karandikar2021soft}.

%Here, $H_i$ is the entropy of the probability distribution of the $i^{th}$ token ${w_i}$ given the previous tokens $w_{0:i-1}$ in the sequence, ${\overline{w_i}}$ is the ground-truth reference token, $C := \{ i \mid w_i = \overline{{w}_i}\}$ is a set of indices corresponding to correctly predicted tokens, $\widetilde{C} := \{ i \mid w_i \neq \overline{w_i}\}$ is a set of indices corresponding to incorrectly predicted tokens, V  is the size of the vocabulary, $w_{i}^{j}$ is the $j^{th}$ token in the vocabulary, and $P_{\theta}(w_{i}^{j} | w_{0: i-1})$ is the predicted probability of the $j^{th}$ token in the vocabulary. The hyperbolic tangent function is employed to scale the token entropy values, ensuring that  $\tanh(H_i$) lies in the interval [0, 1]. We used the hyperbolic tangent function for its smooth gradient properties following prior works in uncertainty calibration \citep{krishnan2020improving, karandikar2021soft}.

The loss function in Equation~\ref{eqn:uaclm_loss} offers theoretical guarantees as an optimization objective and satisfies the desideratum, converging to a perfect value of zero when all correctly generated tokens have a predictive probability of 1 (indicating high confidence) and scaled predictive entropy of 0 (implying low uncertainty), while all incorrectly generated tokens have a predictive probability of 0 (representing low confidence) and scaled predictive entropy of 1 (implying high uncertainty). 
The differentiable utility functions in the UA-CLM loss as shown in Equation~\ref{eqn:uaclm_loss} steers the predictive probabilities and uncertainty estimates to align with the accuracy of subsequent token predictions in autoregressive models. When the uncertainty estimates in the predicted tokens are misaligned, the loss increases, thereby directing the stochastic gradient computations to drive the loss towards minimization. This loss reduces when the uncertainty estimates conform to the desideratum, enabling the model to produce well-calibrated uncertainties while maximizing accuracy. We present an analysis in Appendix \ref{apdx:loss convergence} illustrating the proposed UA-CLM objective improves the reliability of uncertainty estimates associated with correct and incorrect generated tokens as the loss converges, as compared to fine-tuning with standard CLM loss.

Our proposed UA-CLM loss is designed to be agnostic to various parameter-efficient fine-tuning methods in LLMs; however, in this paper, we leverage and illustrate its application through Low-Rank Adaptation (LoRA)~\citep{lora2022} as described in Algorithm~\ref{alg:ua-clm}. By incorporating uncertainty directly into the loss function, the model can not only learn to improve accuracy but also to understand a meaningful representation of uncertainty in its predictions. This dual emphasis on accuracy and uncertainty in the optimization objective ensures that the model's uncertainty estimates are closely aligned with the actual predictive accuracy of the generated tokens, leading to improved uncertainty calibration in natural language generation.

\begin{figure*}[ht]
\centering
\includegraphics[width=1.0\textwidth]{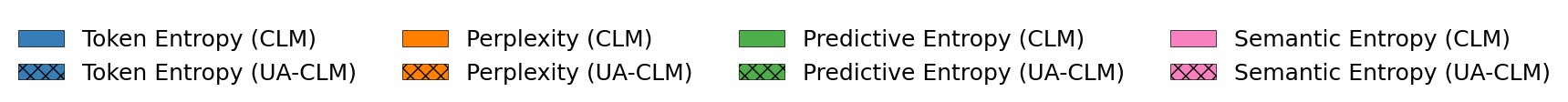}
\includegraphics[width=1.0\textwidth]{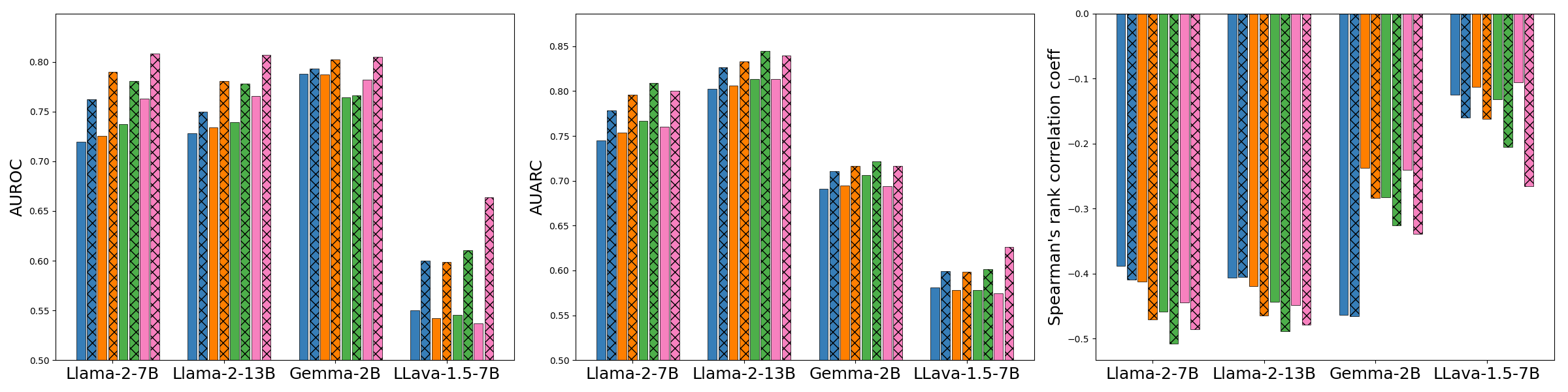}\\
{\small \quad \quad (a) Hallucination detection $\uparrow$\quad\quad\quad (b) Selective generation $\uparrow$\quad\quad\quad (c) Spearman's rank correlation $\downarrow$}
\caption{\small The proposed Uncertainty-aware Causal Language Modeling (UA-CLM)  outperforms standard Causal Language Modeling (CLM) in all four distinct UQ metrics across various models. Performance is evaluated using AUROC for hallucination detection, AUARC for selective generation, and Spearman's rank correlation coefficient $\rho$ for calibration analysis. Higher AUROC, AUARC, and lower Spearman's $\rho$ (stronger inverse correlation between text quality and uncertainty estimate) is desired.}
\label{fig:bar_plot}
\end{figure*}

\begin{algorithm}[ht]
\small
\caption{\small{Uncertainty-aware CLM fine-tuning in LLMs}}
\label{alg:ua-clm}
%\begin{algorithmic}[1]
\begin{algorithmic}
\State \textbf{Input:} Pre-trained LLM $\mathbb{M}$ with parameters $\phi$, LoRA parameters $\theta$, learning rate $\eta$, epochs $E$, training data $\mathcal{D}$
\State \textbf{Output:} Uncertainty calibrated fine-tuned LLM $\mathbb{M}'$
\State Initialize LoRA parameters $\theta$, and freeze $\phi$ 
\For{$epoch = 1$ to $E$}
    \For{each batch $B \subseteq \mathcal{D}$}
        \State Model forward pass to get $P_{\phi+\theta}(w_i | w_{0: i\hbox{-}1})$ and token uncertainty estimate ${H}_i$ 
        \State Compute UA-CLM loss $\mathcal{L^{\phi+\theta}_{\mathrm{UA\hbox{-}CLM}}}$ %\Comment{Equation~\ref{eqn:uaclm_loss}}
        \State Compute gradients of loss function w.r.t. $\theta$,\\ $\nabla_{\theta}\mathcal{L^{\phi+\theta}_{\mathrm{UA\hbox{-}CLM}}}$
        \State Update LoRA parameters: \\$\theta \leftarrow \theta - \eta \cdot \nabla_{\theta}\mathcal{L^{\phi+\theta}_{\mathrm{UA\hbox{-}CLM}}}$
    \EndFor
\EndFor
\State $\theta^* \leftarrow \theta$
\State $\mathbb{M}' \gets \mathbb{M}$ with updated LoRA parameters $\theta^*$
\State \textbf{return} $\mathbb{M}'$
\end{algorithmic}
\end{algorithm}

\section{Experiments and Results}

We perform extensive empirical evaluation to compare our proposed uncertainty-aware causal language modeling (UA-CLM) fine-tuning method to the standard causal language modeling (CLM) fine-tuning, pre-trained baseline, UnLikelihood training (ULT) \citep{welleckneural}, and Calibration Tuning (CT) \citep{kapoor2024calibration} methods. We evaluate on free-form natural language generation tasks. Our comprehensive evaluation rigorously assesses the quality of uncertainty estimates and the quality of the generated text. This includes an analysis of broadly four aspects: hallucination detection, uncertainty-guided selective generation, out-of-domain prompt detection, and calibration analysis.% based on the inverse correlation between the uncertainty estimates and the quality of generated text. 

\subsection{Experimental settings}

\paragraph{Datasets.} We utilize several free-form question-answering (QA) datasets: CoQA \citep{reddy2019coqa}, TriviaQA \citep{joshi2017triviaqa}, and OK-VQA \citep{marino2019ok}. CoQA and TriviaQA are commonly used for benchmarking uncertainty quantification in LLMs \cite{kuhn2023semantic, lin2024generating, farquhar2024detecting}, while OK-VQA extends our evaluation to large vision language models (LVLMs). For out-of-domain prompt detection, we use the BioASQ \citep{bioasq2023krithara} dataset, and for long-form text generation, we employ BioGen \citep{min2023factscore}. In our experiments, we use the development split of CoQA (8,000 samples), the validation split of TriviaQA (10,000 samples), and the validation split of OK-VQA (5,000 samples). We allocate 20\% of each dataset for fine-tuning and the remaining 80\% for testing. Additional details on datasets and prompt settings are provided in Appendix \ref{apdx:datasets}.

\paragraph{Models.}
We use the Llama-2 models with 7B and 13B parameters~\citep{llama2023touvron} and the Gemma model with 2B parameters~\citep{gemma2024} for the free-form QA and biography generation experiments. Additionally, we utilize the LLaVA-1.5 model with 7B parameters~\citep{llava2024liu} for the open-ended visual question-answering task. Our method is model-agnostic and can be applied to the latest LLMs and LVLMs.
\begin{table*}[t]
\caption{\small Evaluation of uncertainty quantification: Comparative analysis of the proposed Uncertainty-aware Causal Language Modeling (UA-CLM) with pre-trained baseline, standard Causal Language Modeling (CLM), UnLikelihood Training (ULT) \citep{welleckneural}, and Calibration Tuning (CT) \citep{kapoor2024calibration} methods. The comparison spans different datasets and models, with quality of UQ evaluated using the Area Under the Receiver Operating Characteristic (AUROC) for hallucination detection and the Area Under the Accuracy-Rejection Curve (AUARC) for selective generation based on four different uncertainty metrics. The results are average of three runs from different random seeds. The best values are in \textbf{bold}.}
\resizebox{\textwidth}{!}{%
\begin{tabular}{@{}lllcccccccccc@{}}
\toprule
\multirow{2}{*}{Dataset} & \multirow{2}{*}{Model} & \multirow{2}{*}{\begin{tabular}[c]{@{}l@{}}Finetuning \\ Method\end{tabular}} & \multicolumn{4}{c}{AUROC $\uparrow$ (Hallucination detection)} &  &  & \multicolumn{4}{c}{AUARC $\uparrow$ (Uncertainty-guided selective generation) } \\ \cmidrule(l){4-7} \cmidrule(l){9-13} 
 &  &  & \begin{tabular}[c]{@{}c@{}}Token \\ Entropy\end{tabular} & Perplexity & \begin{tabular}[c]{@{}c@{}}Predictive \\ Entropy\end{tabular} & \begin{tabular}[c]{@{}c@{}}Semantic \\ Entropy\end{tabular} &  &  & \begin{tabular}[c]{@{}c@{}}Token \\ Entropy\end{tabular} & Perplexity & \begin{tabular}[c]{@{}c@{}}Predictive \\ Entropy\end{tabular} & \begin{tabular}[c]{@{}c@{}}Semantic \\ Entropy\end{tabular} \\ \midrule
% &  &  &  &  &  &  &  &  &  &  &  &  \\
\multirow{17}{*}{CoQA} & \multirow{5}{*}{Llama-2-7B} & Pre-trained & 0.5813 & 0.6324 & 0.6686 & 0.7467 &  &  & 0.8361 & 0.8606 & 0.9348 & 0.9411 \\ 
 &  & CLM & 0.6252 & 0.6320 & 0.6635 & 0.6889 &  &  & 0.9435 & 0.9444 & 0.9508 & 0.9530 \\ 
 &  & ULT  & 0.5790 & 0.5915 & 0.6793 & 0.6495 &  &  & 0.9212 & 0.9219 & 0.9315 & 0.9326 \\ 
  &  & CT  & 0.6175 & 0.6571 & 0.6706 & 0.7292 &  &  & 0.8603 & 0.8780 & 0.9029 & 0.9075 \\ 
 &  & UA-CLM & \textbf{0.6955} & \textbf{0.7398} & \textbf{0.7413} & \textbf{0.7741} &  &  & \textbf{0.9603} & \textbf{0.9657} & \textbf{0.9699} & \textbf{0.9716} \\
 \cmidrule(r){2-13}
 %&  &  &  &  &  &  &  &  &  &  &  &  \\
 & \multirow{5}{*}{Llama-2-13B} & Pre-trained & 0.6027 & 0.6404 & 0.6679 & 0.7111 &  &  & 0.8672 & 0.8940 & 0.9137 & 0.9209 \\ 
 &  & CLM & 0.6302 & 0.6348 & 0.6815 & 0.6910 &  &  & 0.9579 & 0.9584 & 0.9659 & 0.9661 \\ 
 &  &  ULT & 0.6323 & 0.6523 & 0.6883 & 0.7158 &  &  & 0.9510 & 0.9534 & 0.9592 & 0.9612 \\ 
  &  &  CT & 0.5299 & 0.5599 & 0.6072 & 0.6958 &  &  & 0.8497 & 0.8647 & 0.8944 & 0.9200 \\ 
 &  & UA-CLM & \textbf{0.6701} & \textbf{0.7255} & \textbf{0.7363} & \textbf{0.7694} &  &  & \textbf{0.9645} & \textbf{0.9700} & \textbf{0.9784} & \textbf{0.9792} \\
 \cmidrule(r){2-13}
 %&  &  &  &  &  &  &  &  &  &  &  &  \\
 & \multirow{4}{*}{Gemma-2B} & Pre-trained & 0.7073 & 0.7089 & 0.6962 & 0.7635 &  &  & 0.9271 & 0.9339 & 0.9235 & 0.9452 \\ 
 &  & CLM & 0.7723 & 0.7606 & 0.7295 & 0.7618 &  &  & 0.9468 & 0.9454 & 0.9619 & 0.9655 \\ 
 &  & ULT & 0.7097 & 0.6921 & 0.6540 & 0.7162 &  &  & 0.9172 & 0.9152 & 0.9093 & 0.9168 \\ 
 &  & UA-CLM & \textbf{0.7780} & \textbf{0.7837} & \textbf{0.7358} & \textbf{0.7871} &  &  & \textbf{0.9652} & \textbf{0.9668} & \textbf{0.9671} & \textbf{0.9672} \\
 \midrule
% &  &  &  &  &  &  &  &  &  &  &  &  \\
\multirow{17}{*}{TriviaQA} & \multirow{5}{*}{Llama-2-7B} & Pre-trained & 0.7687 & 0.8220 & 0.8191 & 0.8315 &  &  & 0.8050 & 0.8259 & 0.8251 & 0.8369 \\ 
 &  & CLM & 0.8135 & 0.8192 & 0.8108 & 0.8371 &  &  & 0.8617 & 0.8615 & 0.8558 & 0.8630 \\ 
 &  & ULT & 0.7676 & 0.8003 & 0.8004 & 0.8276 &  &  & 0.8273 & 0.8418 & 0.8448 & 0.8519 \\ 
  &  & CT & 0.7714 & 0.8211 & 0.8037 & 0.8233 &  &  & 0.8571 & 0.8812 & 0.8769 & 0.8834 \\ 
 &  & UA-CLM & \textbf{0.8293} & \textbf{0.8393} & \textbf{0.8197} & \textbf{0.8423} &  &  & \textbf{0.8879} & \textbf{0.8927} & \textbf{0.8780} & \textbf{0.8934} \\
  \cmidrule(r){2-13}
 & \multirow{5}{*}{Llama-2-13B} & Pre-trained & 0.7984 & 0.8123 & 0.7801 & 0.8365 &  &  & 0.8441 & 0.8574 & 0.8584 & 0.8587 \\ 
 &  & CLM  & 0.8264 & 0.8333 & 0.7971 & 0.8407 &  &  & 0.8798 & 0.8807 & 0.8708 & 0.8829 \\ 
  &  & ULT & 0.8240 & \textbf{0.8485} & \textbf{0.8245} & \textbf{0.8456} &  &  & 0.8949 & 0.9055 & 0.9013 & 0.9063 \\ 
  &  & CT & 0.7338 & 0.7897 & 0.7991 & 0.8222 &  &  & 0.8460 & 0.8914 & 0.9007 & 0.9019 \\ 
 &  & UA-CLM & \textbf{0.8297} & 0.8352 & 0.8033 & 0.8447 &  &  & \textbf{0.9200} & \textbf{0.9254} & \textbf{0.9155} & \textbf{0.9252} \\
  \cmidrule(r){2-13}
 %&  &  &  &  &  &  &  &  &  &  &  &  \\
 & \multirow{4}{*}{Gemma-2B} & Pre-trained & 0.7633 & 0.7719 & 0.7920 & 0.8127 &  &  & 0.6912 & 0.7225 & 0.7127 & 0.7279 \\ 
 &  & CLM & 0.8030 & 0.8138 & \textbf{0.7989} & 0.8018 &  &  & 0.7256 & 0.7251 & 0.7162 & 0.7198 \\ 
 &  & ULT & 0.7935 & 0.8134 & 0.7912 & 0.8035 &  &  & 0.7212 & 0.7429 & 0.7246 & 0.7413 \\ 
 &  & UA-CLM & \textbf{0.8085} & \textbf{0.8211} & 0.7960 & \textbf{0.8228} &  &  & \textbf{0.7373} & \textbf{0.7436} & \textbf{0.7258} & \textbf{0.7453} \\
 \midrule 
% &  &  &  &  &  &  &  &  &  &  &  &  \\
\multirow{2}{*}{OK-VQA} & \multirow{2}{*}{LLaVA-1.5-7B} & CLM & 0.5504 & 0.5419 & 0.5455 & 0.5370 &  &  & 0.5809 & 0.5781 & 0.5790 & 0.5747 \\ [0.01cm]
 &  & UA-CLM & \textbf{0.6001} & \textbf{0.5984} & \textbf{0.6106} & \textbf{0.6638} &  &  & \textbf{0.5989} & \textbf{0.5965} & \textbf{0.6012} & \textbf{0.6265} \\ 
 \bottomrule
\end{tabular}%
}
\label{tab:uncertainty_eval}
\end{table*}

\paragraph{Fine-tuning.} 
We utilize Low-rank Adaptation (LoRA) ~\citep{lora2022} method to perform parameter-efficient fine-tuning with less than 1\% of model parameters. The models undergo fine-tuning as described in Section~\ref{sec:proposed_method} for the uncertainty-aware causal language modeling method. For all our experiments, the models are fine-tuned for a concise duration of 3 epochs. The optimization is carried out using the AdamW optimizer~\citep{loshchilov2017decoupled}, with an initial learning rate of $1e{\hbox{-}4}$, a weight decay of $0.001$, and a warm-up ratio of $0.03$. For comparison with related works, we also fine-tune models using the standard causal language modeling and unlikelihood training loss functions. We follow the same setup for all the methods for a fair comparison. During the fine-tuning process, only the LoRA parameters are updated, while all other model parameters remain frozen. We provide more details on the hyperparameters and implementation in Appendix~\ref{apdx:finetuning}, to facilitate the reproducibility of the results.

\paragraph{UQ metrics.} To assess the uncertainty quantification in free-form text generation, we employ four widely used metrics: mean token entropy~\citep{fomicheva2020unsupervised}, perplexity~\citep{fadeeva2023lm}, predictive entropy~\citep{malininuncertainty} and semantic entropy~\citep{kuhn2023semantic}. The predictive entropy and semantic entropy are estimated by generating 5 stochastic sequences from the model, each obtained through temperature sampling with a temperature setting of T=0.3. 

\begin{figure*}[t]
\centering
\includegraphics[width=0.77\textwidth]{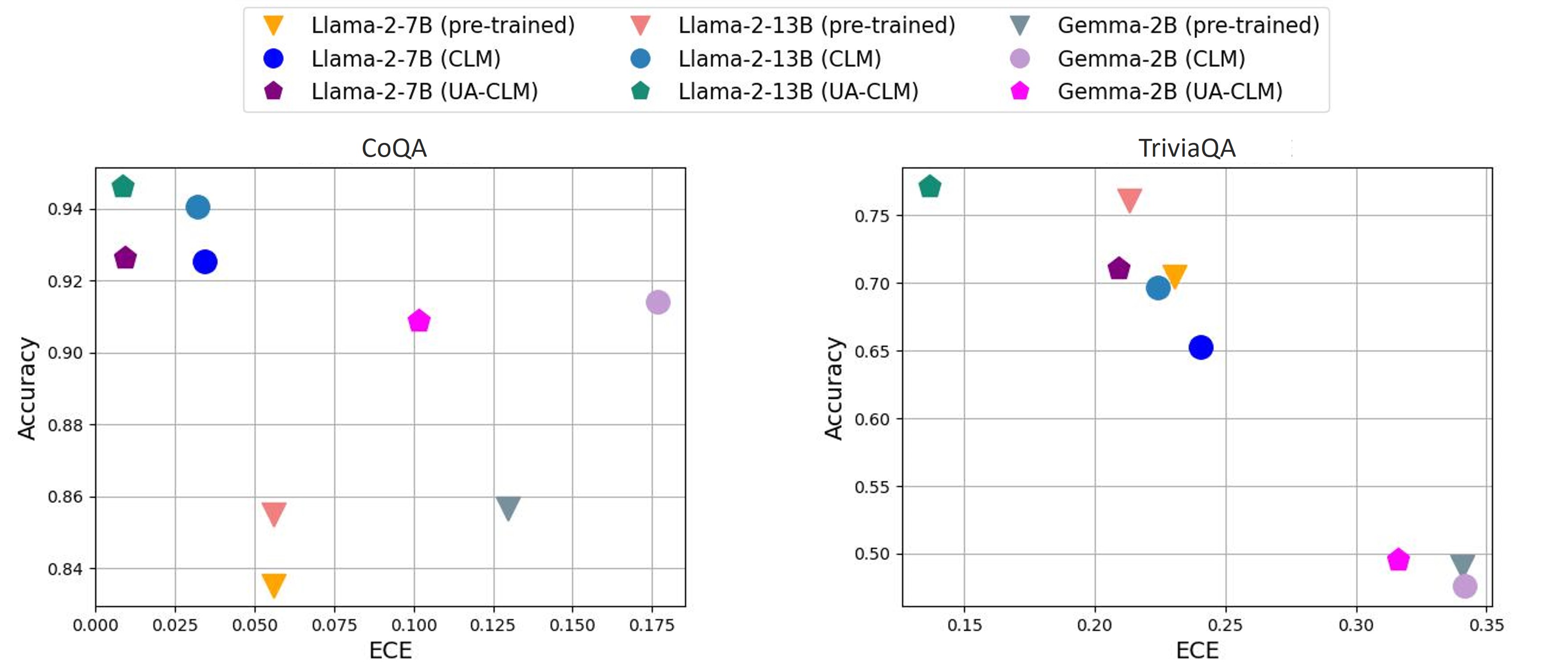}
\caption{\small Accuracy versus Expected Calibration Error (ECE) comparison between UA-CLM, CLM, and pre-trained baseline across different LLM architectures on CoQA and TriviaQA datasets. The ideal model should have high accuracy and low ECE, indicating accurate predictions with well-calibrated uncertainty quantification (upper-left of the plot). The ECE of models fine-tuned with UA-CLM shows significant improvement compared to the pre-trained baseline and CLM fine-tuning.}
\label{fig:acc_vs_ece_main}
\end{figure*}
\begin{table*}[h]
\caption{\small Generated text quality and calibration evaluation: Comparative analysis of Uncertainty-aware Causal Language Modeling (UA-CLM) fine-tuning method with standard Causal Language Modeling (CLM) fine-tuning, pre-trained baseline,  UnLikelihood training (ULT) \citep{welleckneural} and Calibration Tuning (CT) \citep{kapoor2024calibration} methods. The results in the table indicate that UA-CLM achieves higher ROUGE-L and accuracy, and lower expected calibration error (ECE) as compared to other methods.}
\resizebox{\textwidth}{!}{%
\begin{tabular}{@{}lcccccccccccc@{}}
\toprule
\multirow{2}{*}{\begin{tabular}[c]{@{}l@{}}Finetuning   \\ Method\end{tabular}} & \multicolumn{3}{c}{Llama-2-7B (CoQA)} & \multicolumn{3}{c}{Llama-2-7B (TriviaQA)} & \multicolumn{3}{c}{Llama-2-13B (CoQA)} & \multicolumn{3}{c}{Llama-2-13B (TriviaQA)} \\ \cmidrule(l){2-4} \cmidrule(l){5-7} \cmidrule(l){8-10} \cmidrule(l){11-13} 
 & ROUGE-L $\uparrow$ & Accuracy $\uparrow$ & ECE $\downarrow$ & ROUGE-L $\uparrow$ & Accuracy $\uparrow$ & ECE $\downarrow$ & ROUGE-L $\uparrow$ & Accuracy $\uparrow$ & ECE $\downarrow$ & ROUGE-L $\uparrow$ & Accuracy $\uparrow$ & ECE $\downarrow$ \\ \cmidrule(r){1-13}
Pre-trained & 0.7449 & 0.8350 & 0.0561 & 0.6654 & 0.7048 & 0.2304 & 0.7832 & 0.855 & 0.0559 & 0.7160 & 0.7610 & 0.2133 \\ [0.05cm]
CLM & \textbf{0.8886} & 0.9253 & 0.0343 & 0.6037 & 0.6529 & 0.2407 & 0.9106 & 0.9406 & 0.0323 & 0.6588 & 0.6967 & 0.2241 \\ [0.05cm]
ULT & 0.8409 & 0.8950 & 0.0588 & 0.6121 & 0.6586 & 0.3111 & 0.8771 & 0.925 & 0.0595 & 0.6875 & 0.7309 & 0.1517 \\ [0.05cm]
CT & 0.7437 & 0.8125 & 0.0410 & 0.6600 & 0.6987 & 0.2276 & 0.8022 & 0.8725 & 0.0992 & 0.7018 & 0.7429 & 0.1937 \\ [0.05cm]
UA-CLM & \textbf{0.8882} & \textbf{0.9264} & \textbf{0.0094} & \textbf{0.6679} & \textbf{0.7108} & \textbf{0.2090} & \textbf{0.9118} & \textbf{0.9461} & \textbf{0.0084} & \textbf{0.7277} & \textbf{0.7710} & \textbf{0.1365} \\ \bottomrule
\end{tabular}%
}
\label{tab:text_quality}
\end{table*}

\subsection{Evaluation and Results}
\paragraph{Hallucination detection. }
We assessed the ability to detect hallucinations in generated text using uncertainty estimates, employing the Area Under the Receiver Operating Characteristic (AUROC)~\citep{davis2006relationship} as a measure of the model's effectiveness in identifying correct responses and flagging hallucinations, following ~\citep{farquhar2024detecting}. A higher AUROC indicates better performance in detecting hallucinations. The bar plot depicted in Figure \ref{fig:bar_plot}(a) provides a visual representation of AUROC performance for both CLM and UA-CLM methods. It shows the improvements in various uncertainty quantification metrics across distinct LLMs and LVLM, highlighting the enhanced reliability of uncertainty estimates in the generated text achieved by UA-CLM. This plot consolidates AUROC metrics from the CoQA and TriviaQA datasets for LLMs and the OK-VQA dataset for LVLM, with specific numbers for corresponding datasets and models provided in Table \ref{tab:uncertainty_eval}.
Our findings show that the UA-CLM method significantly enhances hallucination detection, with improvements of up to 17.1\% on QA tasks and 23.6\% on VQA task compared to the standard CLM method, outperforming ULT and CT methods. UA-CLM achieves better-calibrated uncertainty estimates without compromising text quality or overall accuracy, as shown in Table \ref{tab:text_quality}. For details on text quality metrics and additional results, please refer to Appendix \ref{apdx:text_eval_metrics} and \ref{apdx:additional_results}.

\paragraph{Uncertainty-guided selective generation.}
The ability of LLMs to decide when to generate a response and when to abstain based on uncertainty estimates is crucial for developing trustworthy generative AI models. This capability allows models to recognize and communicate their limitations, which is vital in scenarios where incorrect responses could have serious consequences, such as in medical diagnosis and legal advice. We adopt the methodology from \cite{farquhar2024detecting} and use the Area Under the Accuracy-Rejection Curve (AUARC)~\citep{nadeem2009accuracy} to evaluate decision-making informed by uncertainty estimates in selective generation~\citep{ren2022out}. AUARC is a valuable metric for assessing the quality of a model's uncertainty estimates and its ability to decide when to provide response and when to abstain due to high uncertainty. As shown in Figure \ref{fig:bar_plot}(b) and Table \ref{tab:uncertainty_eval}, UA-CLM outperforms other methods in AUARC scores across datasets and models, indicating better downstream decision-making.

\begin{table*}[]
\caption{\small Evaluating generalization ability of uncertainty calibration on short-form QA task and long-form biography generation task.}
\label{tab:generalization}
\resizebox{\textwidth}{!}{%
\begin{tabular}{@{}lccccccccclcccc@{}}
\toprule
\multirow{4}{*}{Method} & \multicolumn{4}{c}{CoQA \textbf{$\longrightarrow$} TriviaQA} & \multicolumn{1}{l}{} & \multicolumn{4}{c}{TriviaQA \textbf{$\longrightarrow$} CoQA} &  & \multicolumn{4}{c}{CoQA \textbf{$\longrightarrow$} BioGen (Long-form generation)} \\ \cmidrule(l){2-5} \cmidrule(l){7-10} \cmidrule(l){12-15} 
 & \multicolumn{4}{c}{AUROC ↑ (Hallucination detection)} & \multicolumn{1}{l}{} & \multicolumn{4}{c}{AUROC ↑ (Hallucination detection)} &  & \multicolumn{4}{c}{AUROC ↑ (Hallucination detection)} \\ \cmidrule(l){2-5} \cmidrule(l){7-10} \cmidrule(l){12-15} 
 & \begin{tabular}[c]{@{}c@{}}Token   \\ Entropy\end{tabular} & Perplexity & \begin{tabular}[c]{@{}c@{}}Predictive \\ Entropy\end{tabular} & \begin{tabular}[c]{@{}c@{}}Semantic \\ Entropy\end{tabular} &  & \begin{tabular}[c]{@{}c@{}}Token \\ Entropy\end{tabular} & Perplexity & \begin{tabular}[c]{@{}c@{}}Predictive \\ Entropy\end{tabular} & \begin{tabular}[c]{@{}c@{}}Semantic\\  Entropy\end{tabular} & \multicolumn{1}{c}{} & \begin{tabular}[c]{@{}c@{}}Token \\ Entropy\end{tabular} & Perplexity & \begin{tabular}[c]{@{}c@{}}Predictive \\ Entropy\end{tabular} & \begin{tabular}[c]{@{}c@{}}Semantic \\ Entropy\end{tabular} \\ \midrule
CLM & 0.7201 & 0.7847 & 0.7532 & 0.7874 &  & 0.5952 & 0.6349 & 0.6665 & 0.7146 & \multicolumn{1}{c}{} & 0.5653 & 0.5793 & 0.5122 & 0.5281 \\
UA-CLM & \textbf{0.8271} & \textbf{0.8261} & \textbf{0.788} & \textbf{0.8146} &  & \textbf{0.6456} & \textbf{0.6824} & \textbf{0.7154} & \textbf{0.7528} & \multicolumn{1}{c}{} & \textbf{0.6135} & \textbf{0.6123} & \textbf{0.7374} & \textbf{0.5354} \\ \bottomrule
\end{tabular}%
}
\end{table*}

\begin{table}[h!]
\caption{\small Out-of-domain detection: Evaluation with Biomedical question answering (BioASQ) as out-of-domain dataset on Llama-2-7B finetuned with CoQA dataset. The table shows the comparison of CLM and UA-CLM with AUROC and AUPR scores for out-of-domain detection using different uncertainty metrics.}
%\resizebox{0.5\textwidth}{!}{%
\begin{adjustbox}{width=1.0\textwidth}
\begin{tabular}{@{}lcccclcccc@{}}
\toprule
\multirow{2}{*}{Method} & \multicolumn{4}{c}{AUROC $\uparrow$ (Out-of-domain detection)} &  & \multicolumn{4}{c}{AUPR $\uparrow$ (Out-of-domain detection)} \\ \cmidrule(l){2-5} \cmidrule(l){7-10} 
 & \begin{tabular}[c]{@{}c@{}}Token \\ Entropy\end{tabular} & Perplexity & \begin{tabular}[c]{@{}c@{}}Predictive\\  Entropy\end{tabular} & \begin{tabular}[c]{@{}c@{}}Semantic \\ Entropy\end{tabular} &  & \begin{tabular}[c]{@{}c@{}}Token \\ Entropy\end{tabular} & Perplexity & \begin{tabular}[c]{@{}c@{}}Predictive \\ Entropy\end{tabular} & \begin{tabular}[c]{@{}c@{}}Semantic \\ Entropy\end{tabular} \\ \midrule
CLM & 0.7828 & 0.7661 & 0.7500 & 0.7902 &  & 0.8124 & 0.7928 & 0.8570 & 0.8422 \\
UA-CLM & \textbf{0.9025} & \textbf{0.8931} & \textbf{0.7763} & \textbf{0.9061} & \textbf{} & \textbf{0.9235} & \textbf{0.9117} & \textbf{0.8958} & \textbf{0.9250} \\ \bottomrule
\end{tabular}%
%}
\end{adjustbox}
\label{tab:ood_detection}
\end{table}

\paragraph{Uncertainty calibration analysis.}
Calibration is crucial for ensuring the quality of uncertainty estimates. We estimate the sentence-level calibration using the Expected Calibration Error (ECE)~\citep{naeini2015obtaining}  based on the correctness of generated responses. As shown in Table \ref{tab:text_quality}, UA-CLM fine-tuning yields lower ECE compared to other methods. Figure \ref{fig:acc_vs_ece_main} presents accuracy versus ECE plots for the CoQA and TriviaQA datasets across different models, including pre-trained baselines, CLM, and UA-CLM fine-tuning methods. Additionally, we employ Spearman's rank correlation coefficient~\citep{zwillinger1999crc} to evaluate the reliability of uncertainty estimates and examine how well these estimates align with the quality of the generated text. This approach addresses the challenges posed by the varying token lengths in generative outputs. We examined the inverse correlation between uncertainty estimates and ROUGE-L scores, a recognized metric to assess text quality. Ideally, well-calibrated uncertainty estimates should show a negative correlation with text quality: uncertainty should increase as ROUGE-L scores decrease, and vice versa. Figure \ref{fig:bar_plot}(c) illustrates that UA-CLM exhibits a stronger inverse correlation between uncertainty estimates and text quality compared to standard fine-tuning. These results demonstrate the effectiveness of uncertainty-aware fine-tuning to obtain better calibrated uncertainty in free-form text generation tasks.

\paragraph{Generalization and long-form text generation.}
We investigate the generalizability of the proposed UA-CLM method to improve uncertainty calibration.  Since we fine-tune the model in a causal language modeling setup that involves next-token prediction, the learning should be transferable, encouraging generalization. We evaluated Llama-2-7B model fine-tuned with CoQA dataset on TriviaQA dataset, and vice-versa. Additionally, to evaluate the method beyond short-form QA tasks, we assessed its performance on biography generation, a long-form paragraph-level generation task. The Llama-2-7B models fine-tuned on CoQA were given prompts to write biographies of popular figures, whose names are sourced from BioGen \citep{min2023factscore}. For long-form generation, the responses were compared against those obtained from GPT-4 \citep{achiam2023gpt} using same prompts, which served as the ground truth reference. The results in Table \ref{tab:generalization} demonstrate that UA-CLM achieves better generalization in offering reliable uncertainty estimates, as indicated by hallucination detection AUROC. 
\paragraph{Out-of-domain detection.}
We evaluate the model's ability to determine if a given prompt is out-of-domain, meaning it pertains to a query that falls outside the scope of the model's trained knowledge base. The Biomedical question-answering (BioASQ)~\citep{bioasq2023krithara} dataset serves as our out-of-domain dataset, while the CoQA dataset is used to represent in-domain data. We leverage uncertainty estimates to detect out-of-domain prompts and employ two widely recognized metrics for evaluation: AUROC and the Area Under the Precision-Recall curve (AUPR)~\citep{saito2015precision}. The results presented in Table \ref{tab:ood_detection}, demonstrate that UA-CLM significantly outperforms the standard CLM in out-of-domain detection task. This performance is consistently observed across all four uncertainty metrics, with up to 16.5\% improvement in AUROC and up to 15\% improvement in AUPR scores. 

\section{Discussion}
We introduced a novel fine-tuning approach aimed at improving uncertainty calibration in LLMs for natural language generation. We proposed a differentiable uncertainty-aware causal language modeling loss, grounded in decision theory. Our method enhances the model's ability to provide well-calibrated uncertainty estimates, a crucial aspect of trustworthy AI models. Through extensive empirical evaluations on free-form question-answering tasks, we demonstrated that our approach significantly improves the quality of various uncertainty quantification metrics. This improvement enhances the model's ability to detect hallucinations, identify out-of-domain prompts, and make selective generation decisions. Furthermore, we showcased the generalizability of our fine-tuning method in the causal language modeling setup to different text generation tasks. 

Although our method focuses on calibrating token-level uncertainty, experimental results suggest that it also benefits sentence-level semantic uncertainty estimates. We hope this work sets the stage for the exploration of calibrating sentence-level uncertainty and inspires further research to advance uncertainty calibration in LLMs.

In conclusion, this work contributes to the broader goal of developing trustworthy LLMs. The ability to recognize out-of-domain prompts and to acknowledge the limits of a model's pre-trained knowledge through reliable uncertainty quantification paves the way for mitigating hallucinations and enhancing decision-making in AI systems.

\newpage

%\section*{Limitations}

%Our method is demonstrated in white-box model settings, where access to model internals are available for uncertainty-calibrated fine-tuning. In black-box scenarios, where direct access to model internals is restricted, alternative strategies such as prompt tuning or developing an auxiliary model for calibrated uncertainty quantification could be explored. These limitations highlight areas for future exploration and development to broaden its applicability and impact.

%\section*{Ethics Statement}
%The broader impact of our work is its potential to advance the reliability and trustworthiness of AI models, especially in the context of natural language generation. Our method aims to enable the large language models (LLMs) and large vision-language models (VLMs) to generate well-calibrated uncertainty estimates associated with generated responses. This advancement is crucial for developing AI systems that can responsibly abstain from generating incorrect or misleading responses when faced with queries outside their pre-trained knowledge base. This capability is essential for mitigating potential hallucinations, especially in critical domains such as healthcare, legal advice, and education. Reliable uncertainty estimation fosters greater trust in AI systems by ensuring that users are informed about the limitations of the model's generated outputs. Our work contributes to the development of safer LLMs and the more effective deployment of AI technologies, benefiting society through advancements in trustworthy machine learning.

\bibliography{iclr2026_conference}
\bibliographystyle{iclr2026_conference}

\clearpage
\appendix
\section*{Appendix}
%\twocolumn[
%\begin{center}
\section{Loss convergence and Uncertainty estimate analysis during fine-tuning}
%\end{center}
%]
\label{apdx:loss convergence}
\begin{figure*}[htbp]
  \centering
%\begin{subfig}
    \centering
    \includegraphics[width=6.5cm]{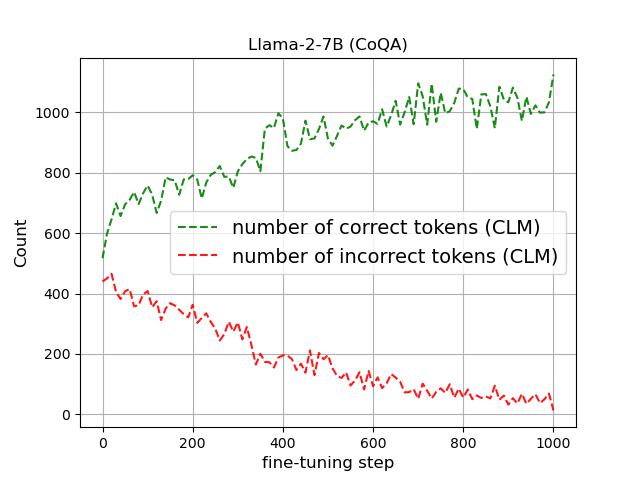}
    \includegraphics[width=6.5cm]{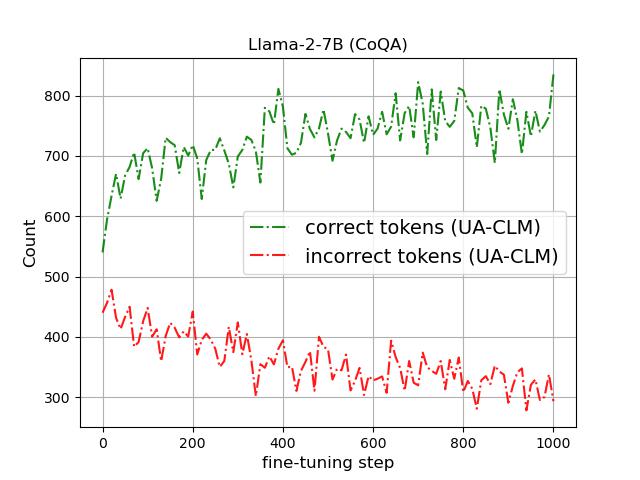}
    
    \small {(a) CLM \quad\quad\quad\quad\quad\quad\quad\quad\quad\quad\quad\quad\quad\quad\quad\quad\quad\quad  (b) UA-CLM}
    \caption{\small Analysis of Correct and Incorrect Token Counts in mini-batch during fine-tuning with CLM and UA-CLM. Both CLM and UA-CLM show increase in correct tokens and a decrease in incorrect tokens as fine-tuning progresses.}%
    \label{apdx_fig:analysis_token_count}
%\end{subfig}
%\begin{subfig}
    \centering
    \includegraphics[width=6.5cm]{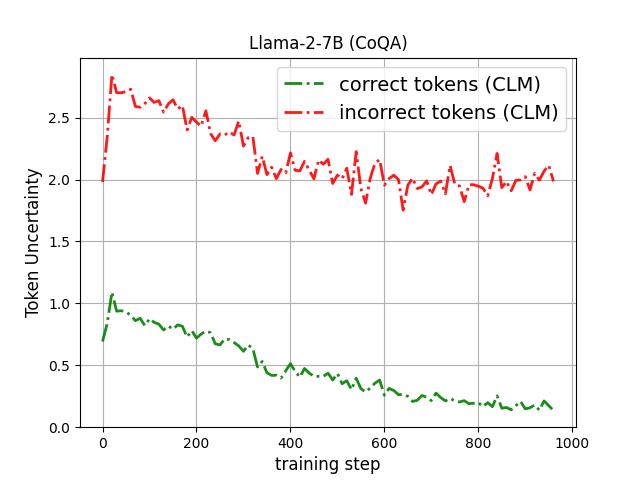}
    \qquad
    \includegraphics[width=6.5cm]{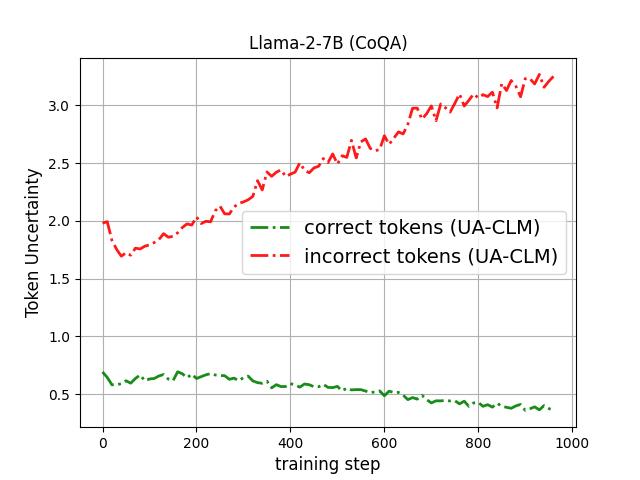}
    
    \small {(a) CLM \quad\quad\quad\quad\quad\quad\quad\quad\quad\quad\quad\quad\quad\quad\quad\quad\quad\quad  (b) UA-CLM}
    \caption{\small Analysis of Token Uncertainty associated with Correct and Incorrect tokens in the mini-batch during fine-tuning with CLM and UA-CLM. A well-calibrated model should provide low uncertainty for correct tokens and higher uncertainty for incorrect tokens. With standard CLM Loss, uncertainty for both correct and incorrect tokens decreases, indicating overconfidence even on incorrect tokens. In contract, with UA-CLM, the uncertainty for incorrect tokens increases and the decreasing uncertainty on correct tokens, supporting that the fine-tuning with UA-CLM improves the reliability of uncertainty estimates.}%
    \label{apdx_fig:analysis_token_unc}
%\end{subfig}
%\begin{subfig}
    \centering
    \includegraphics[width=6.5cm]{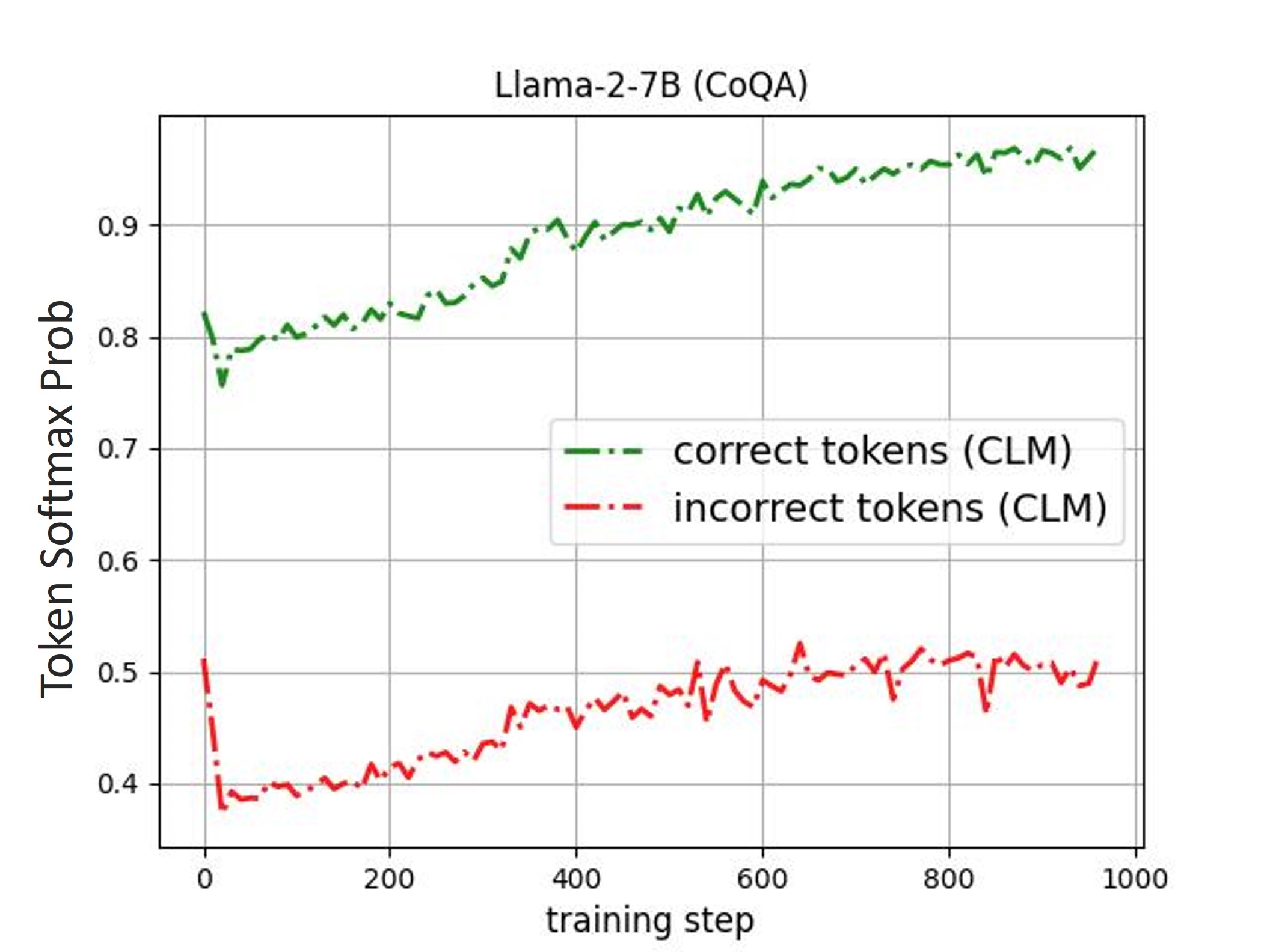}
    \qquad
    \includegraphics[width=6.5cm]{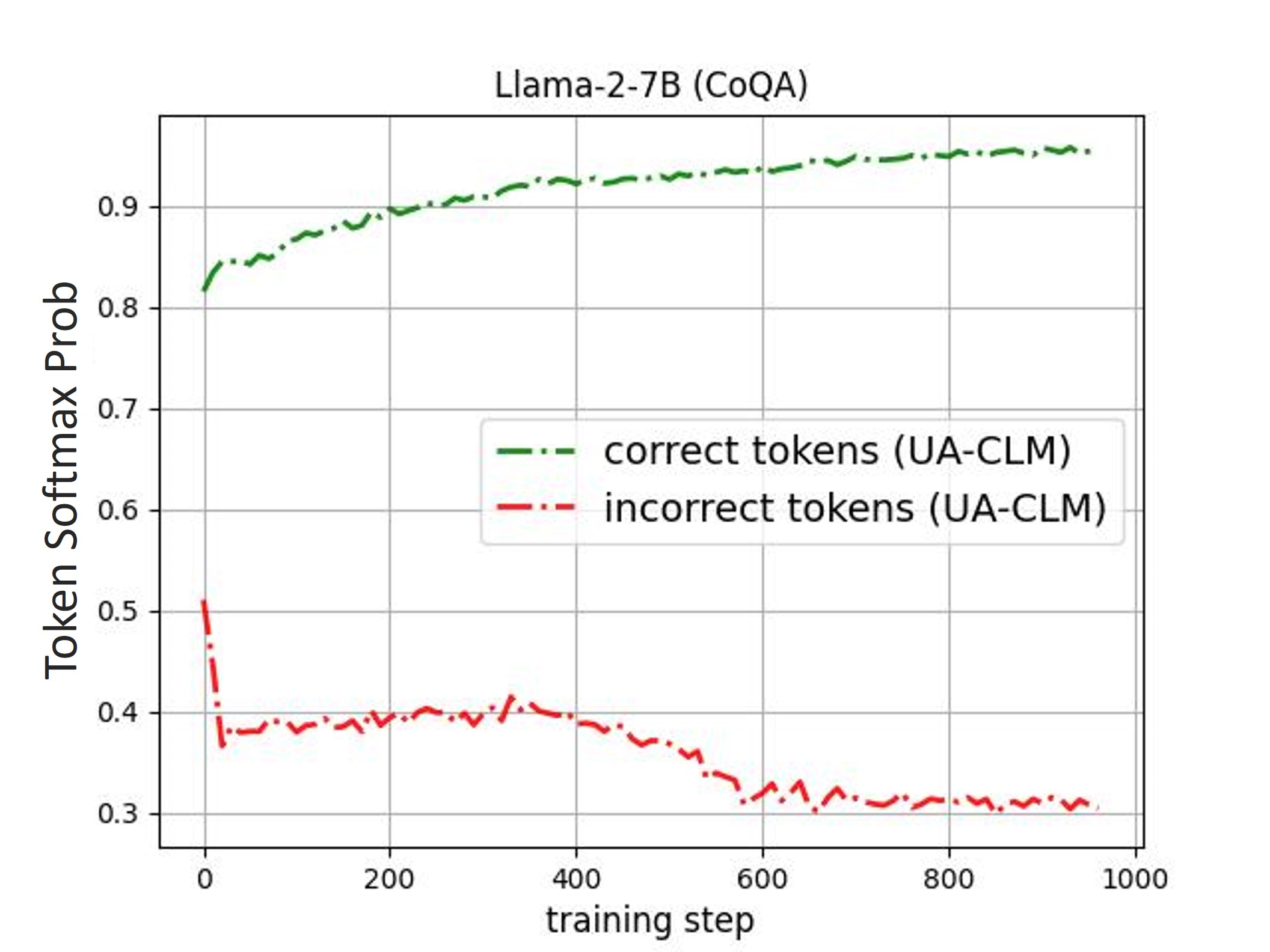}
    
    \small {(a) CLM \quad\quad\quad\quad\quad\quad\quad\quad\quad\quad\quad\quad\quad\quad\quad\quad\quad\quad  (b) UA-CLM}
    \caption{\small Analysis of Token Softmax Probability associated with Correct and Incorrect tokens during fine-tuning with CLM and UA-CLM. A well-calibrated model should assign high probability to correct tokens and lower probability to incorrect tokens. With standard CLM loss, probabilities for both correct and incorrect tokens increase as fine-tuning progress, indicating overconfidence. In contrast, UA-CLM fine-tuning results in higher probabilities for correct tokens and lower probabilities for incorrect tokens, enhancing the reliability of token probability scores}%
    \label{apdx_fig:analysis_token_rob}
%\end{subfig}
\end{figure*}

\begin{figure*}[htbp]
\centering
%\begin{subfigure}
    \centering
    \includegraphics[width=4.4cm]{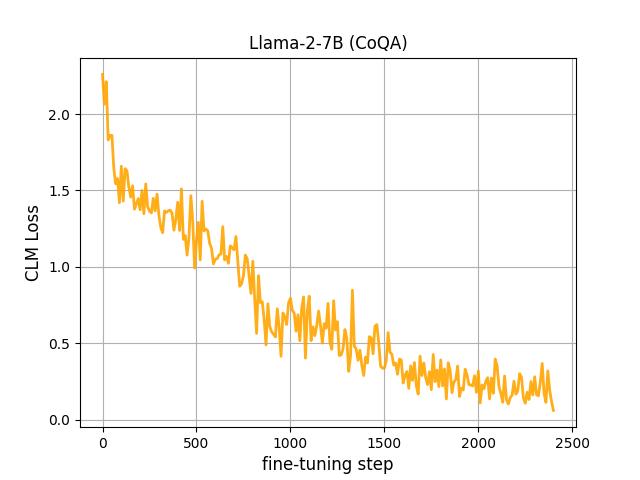}
    \includegraphics[width=4.4cm]{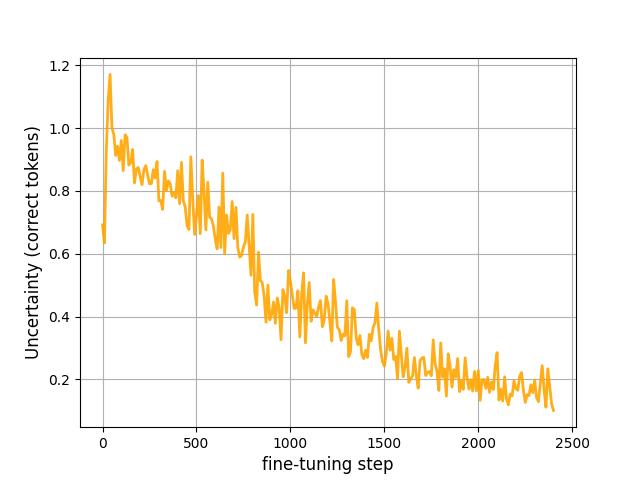}
    \includegraphics[width=4.4cm]
    {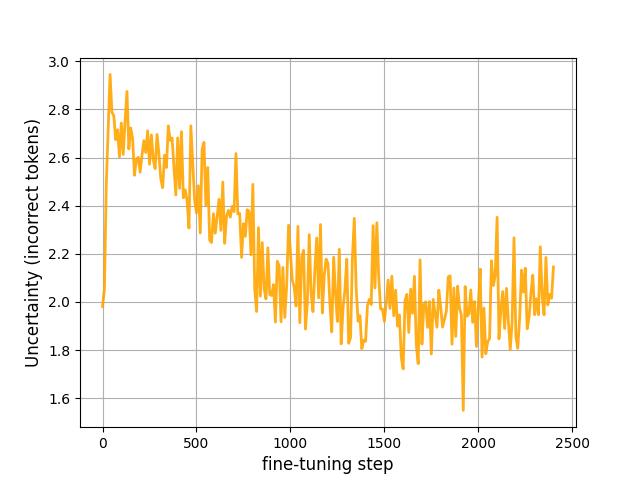}
    \quad\quad\quad\quad\quad\quad\quad\quad\quad\quad\quad\quad\quad\quad\quad\quad\quad\quad\quad\quad\quad\quad\quad\quad\quad\quad\quad\quad\quad \small {(i) CLM} \quad 
    
    \includegraphics[width=4.4cm]
    {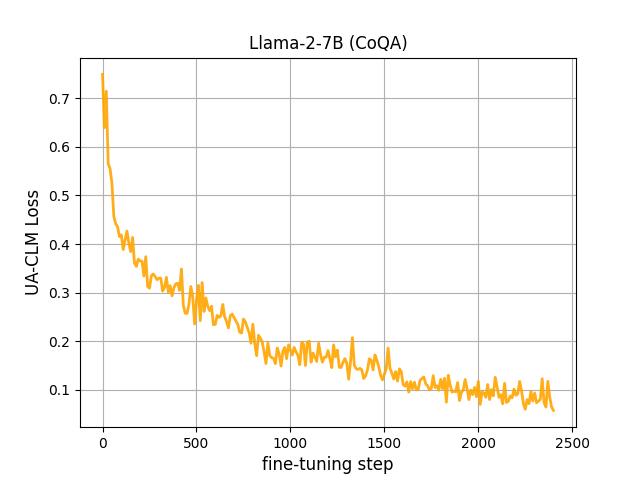}
    \includegraphics[width=4.4cm]{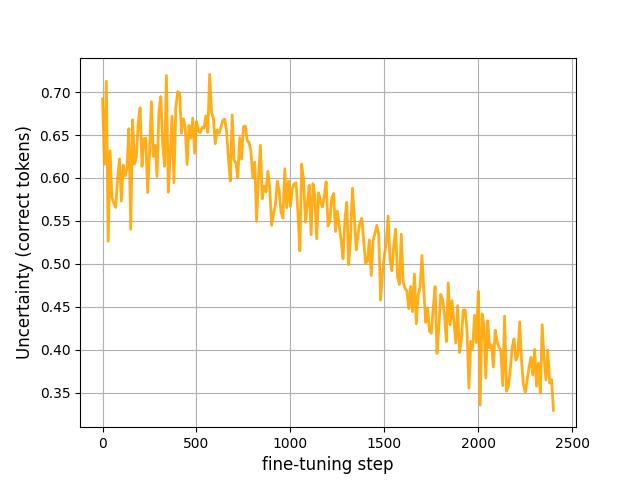}
    \includegraphics[width=4.4cm]{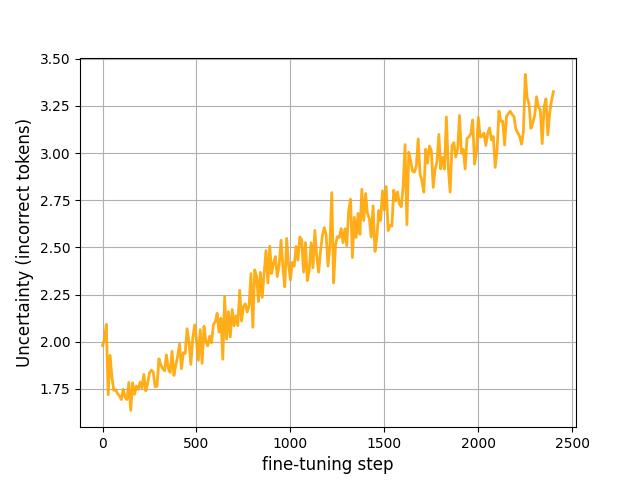}
    \quad\quad\quad\quad\quad\quad\quad\quad\quad\quad\quad\quad\quad\quad\quad\quad\quad\quad\quad\quad\quad\quad\quad\quad\quad\quad\quad\quad\quad \small {(ii) UA-CLM} \quad  
    \caption{\footnotesize Llama-2-7B: Loss convergence and uncertainty values associated with correct and incorrect tokens.}%
    \label{apdx_fig:llama-2-7b_loss_convergence}
%\end{subfigure}

%\begin{subfigure}
    \centering
    \includegraphics[width=4.4cm]{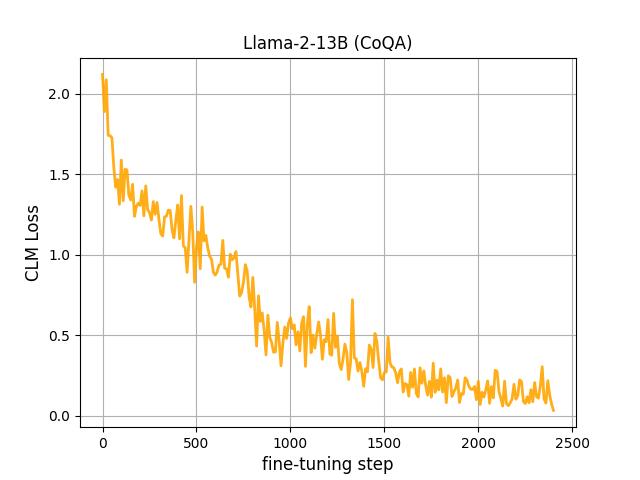}
    \includegraphics[width=4.4cm]{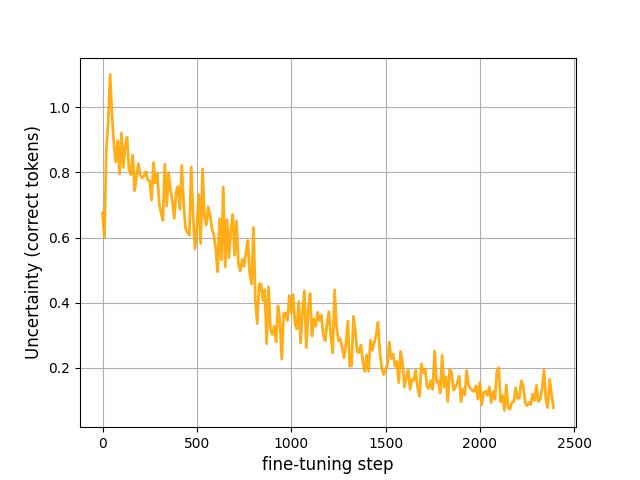}
    \includegraphics[width=4.4cm]{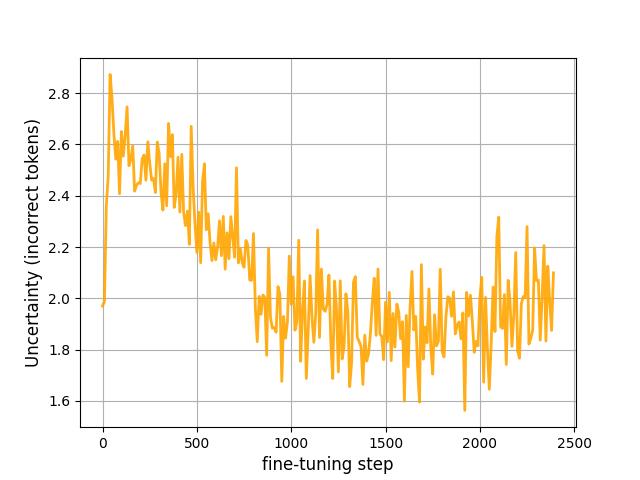}
    
    \quad \small{(i) CLM} 
    
    \includegraphics[width=4.4cm]
    {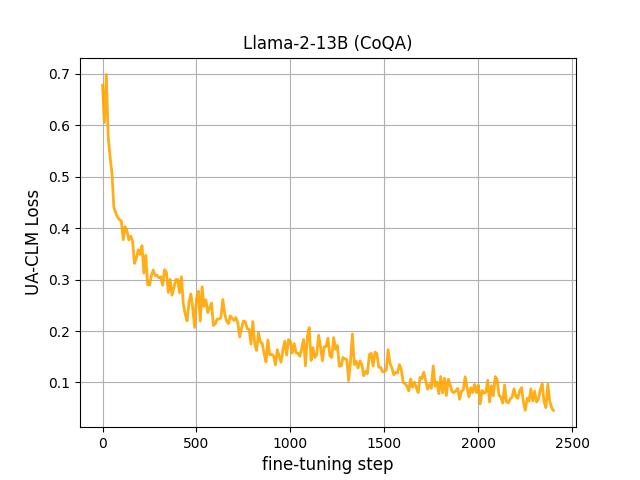}
    \includegraphics[width=4.4cm]{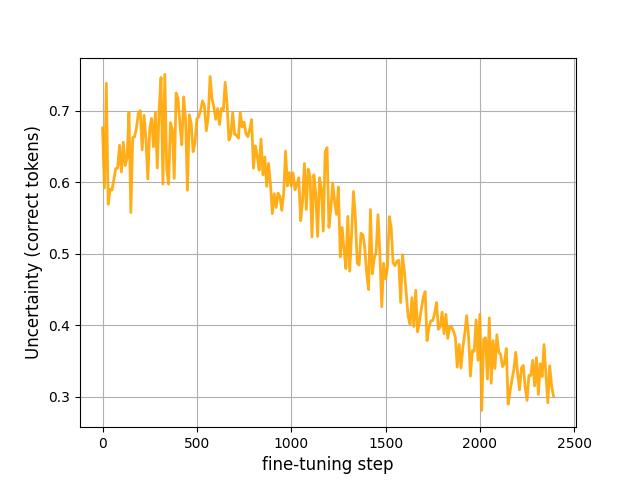}
    \includegraphics[width=4.4cm]{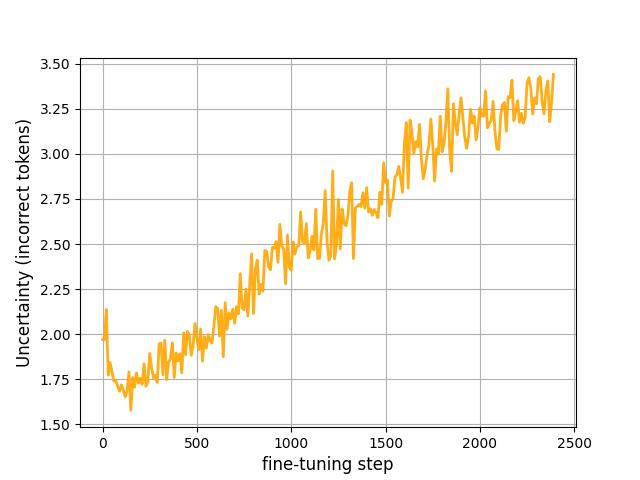}
    
     \quad \small{(ii) UA-CLM} 
    \caption{\footnotesize Llama-2-13B: Loss convergence and uncertainty values for correct and incorrect tokens.}%
    \label{apdx_fig:llama-2-13b_loss_convergence}
%\end{subfigure}

%\begin{subfigure}
    \centering
    \includegraphics[width=13cm]{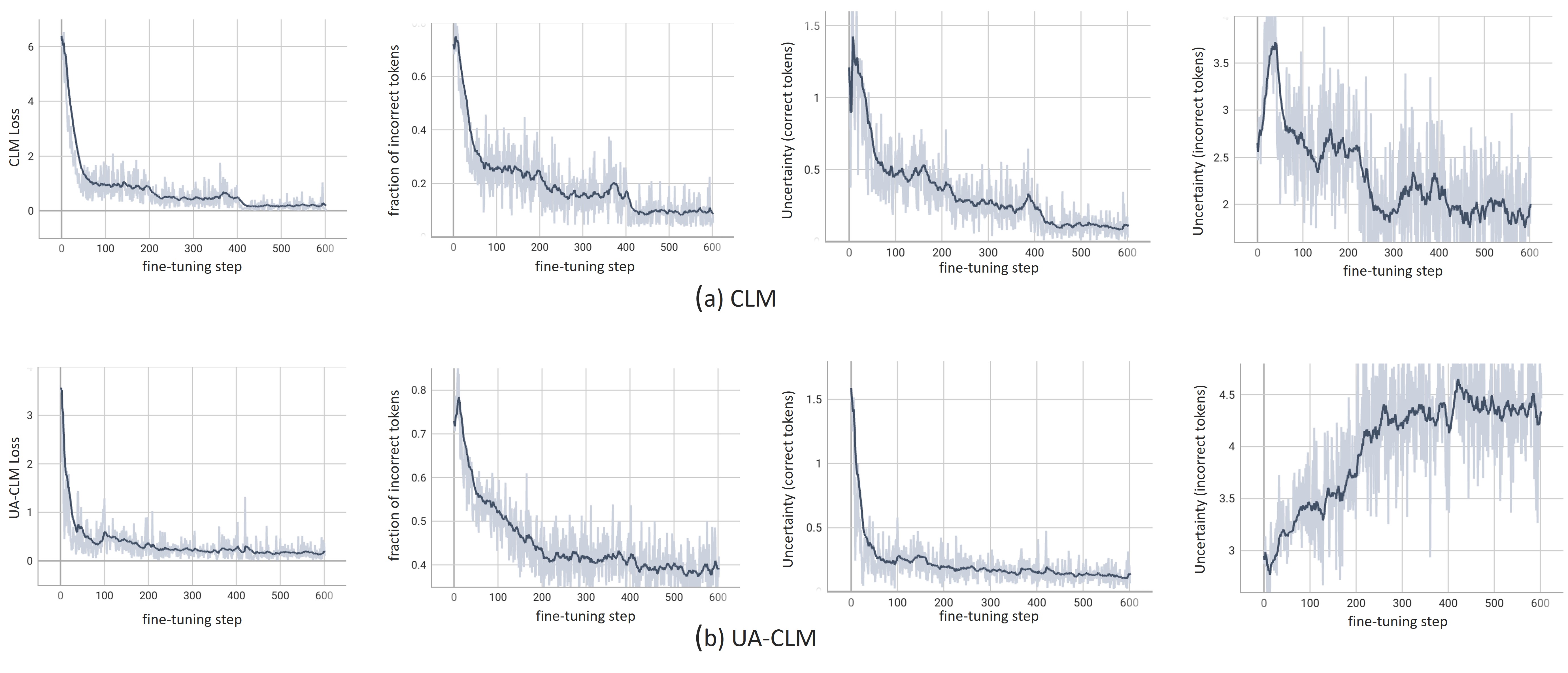}
    \caption{\footnotesize Llava-1.5: Loss convergence and uncertainty values associated with correct and incorrect tokens.}%
    \label{apdx_fig:okvqa_loss_convergence}
%\end{subfigure}
\end{figure*}

We investigate the convergence of fine-tuning and the associated uncertainty values for both correct and incorrect tokens as fine-tuning progress. Figure \ref{apdx_fig:analysis_token_count} shows that both CLM and UA-CLM increase the number of correctly generated tokens while reducing the number of incorrect tokens, thereby improving the accuracy of the generated text.

Regarding uncertainty analysis, as illustrated in Figure \ref{apdx_fig:analysis_token_unc}, the standard CLM loss results in a decrease in uncertainty for both correct and incorrect tokens, indicating overconfidence even when generating incorrect tokens. In contrast, UA-CLM increases the uncertainty for incorrect tokens while decreasing it for correct tokens. These plots suggest that fine-tuning with UA-CLM encourages the model to produce more reliable uncertainty estimates. Additional results for different models and datasets are presented in Figures \ref{apdx_fig:llama-2-7b_loss_convergence}-\ref{apdx_fig:okvqa_loss_convergence}. These findings reveal that our proposed UA-CLM method significantly improves model calibration, supporting the claims of our proposed loss function objectives.

\section{Experimental details}
\subsection{Datasets}
\label{apdx:datasets}
In our study, we utilized a diverse collection of free-form question-answering datasets to evaluate the performance of Large Language Models (LLMs) on phrase-length, sentence-length and paragraph-length text generation tasks. This includes the CoQA~\citep{reddy2019coqa} and TriviaQA~\citep{joshi2017triviaqa} datasets, which are designed to test the model's abilities in text-based question answering, and the BioASQ~\citep{bioasq2023krithara} dataset for out-of-domain evaluation, which presents domain-specific challenges in biomedical field. Additionally, to assess the capabilities of Large Vision Language Models (LVLMs), we utilize the OK-VQA~\citep{marino2019ok} dataset for open-ended visual question-answering tasks. CoQA and BioASQ datasets involve responses that are either sentence-length or phrase-length. TriviaQA and OK-VQA datasets involve phrase-length responses. To evaluate long-form text generation, we use the BioGen dataset \citep{min2023factscore}, which contains names of popular figures for biography generation. We prompt the model to write biographies, to assess on paragraph-length text generation task.

\begin{figure}[ht]
\centering
\includegraphics[width=0.75\textwidth,height=0.4\textheight,keepaspectratio]{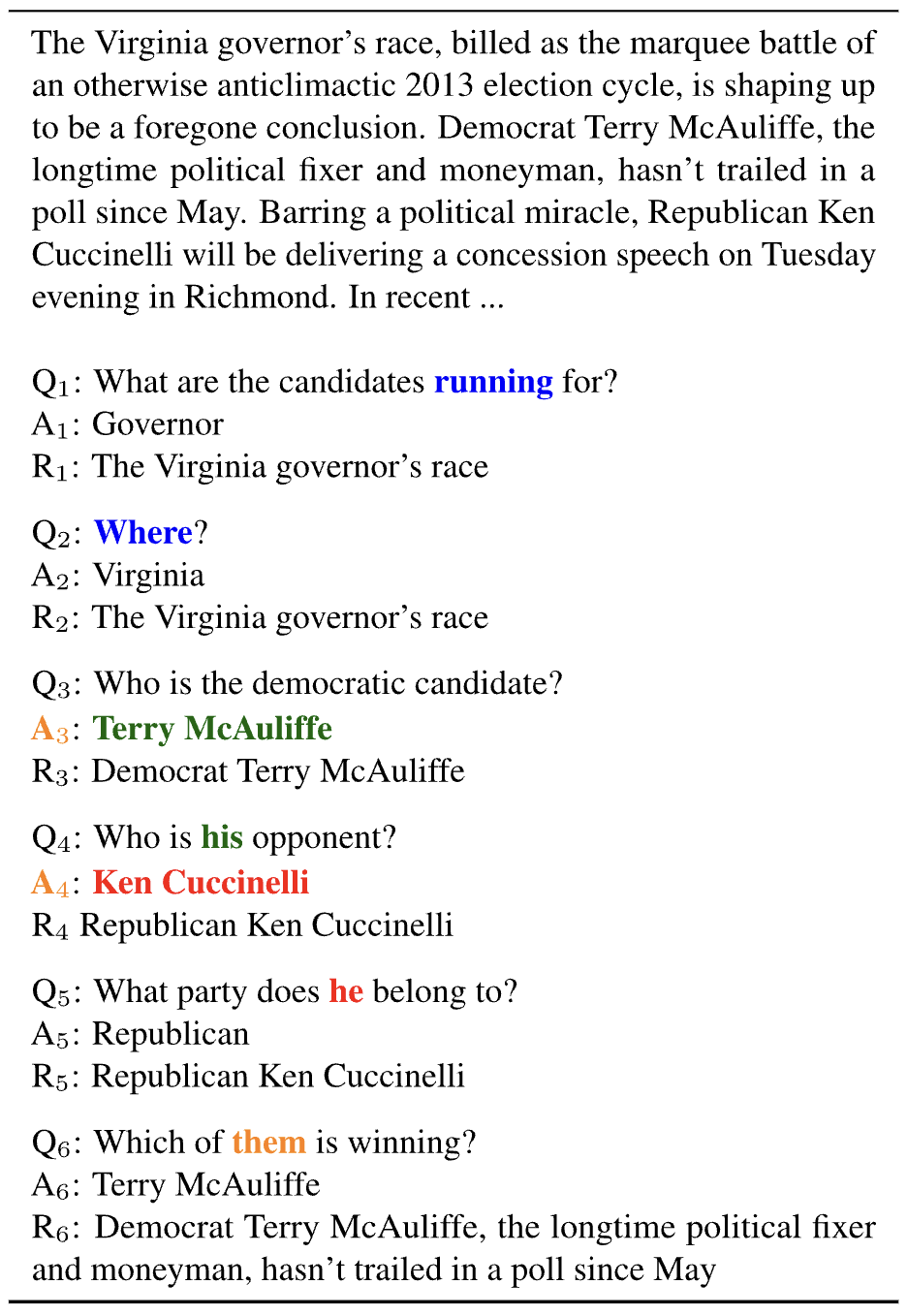}
\caption{Sample from CoQA~\citep{reddy2019coqa} illustrating the co-reference chain of conversational questions.}
\label{coqa:dataset_samples}
\end{figure}

\begin{figure*}[ht]
\centering
\includegraphics[width=1.0\textwidth]{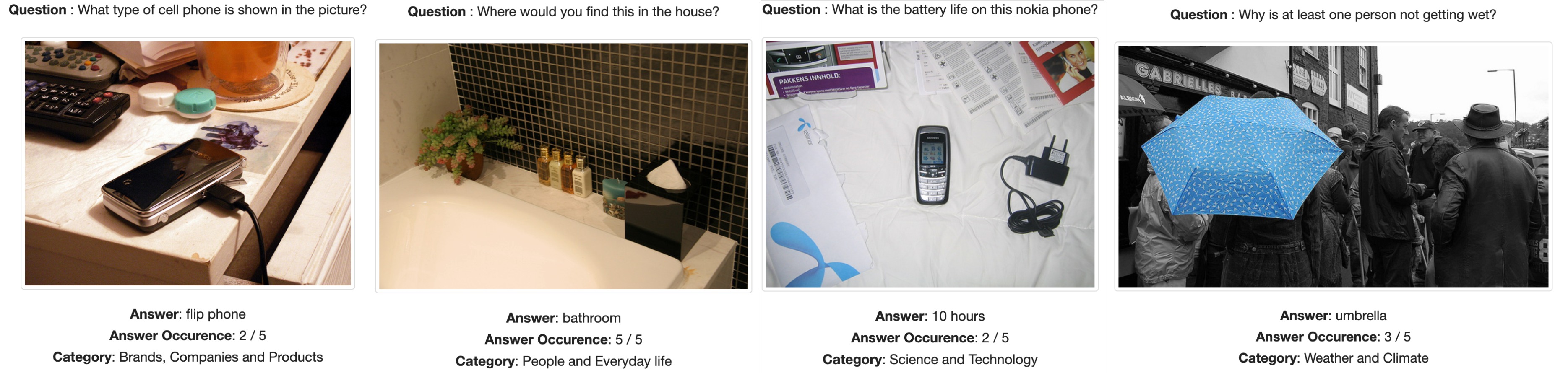}
\caption{Data samples from OK-VQA~\citep{marino2019ok} across different knowledge categories.}
\label{okvqa:dataset_samples}
\end{figure*}

\paragraph{CoQA}
Conversational Question Answering (CoQA)~\citep{reddy2019coqa} dataset was developed to evaluate models' ability to respond to natural, dialogue-based questions, with free-form text answers supported by highlighted evidence from the passage. The full dataset comprises of 127k question-answer pairs derived from 8k conversations based on text passages across 7 distinct domains. For all our experiments, we utilize the development subset of CoQA, which consists of 8k question-answer pairs. Figure \ref{coqa:dataset_samples} shows the color-coded co-reference chains in CoQA as illustrated in the \citep{reddy2019coqa}.
\paragraph{TriviaQA}
TriviaQA~\citep{joshi2017triviaqa} is a reading comprehension dataset consisting of over 650k question-answer-evidence triplets. It includes 95,000 question-answer pairs authored by trivia enthusiasts, along with an average of six independently gathered evidence documents per question, providing high-quality distant supervision for answering the questions. In our experiment, we used the validation split of the dataset with around 10,000 question-answer pairs. Table \ref{tab:triviaqa_samples} shows some of the samples from the dataset.
\paragraph{OK-VQA}
Outside Knowledge-Visual Question Answering benchmarks~\citep{marino2019ok} consists of visual queries where the image content alone is not sufficient to answer the questions. Thus, it requires models to incorporate external knowledge to generate accurate answers. The dataset consists of 14k questions across 10 knowledge categories. In our experiment, we used the validation split of the dataset with around 5k question-answer pairs. Figure \ref{okvqa:dataset_samples} shows a few samples from the dataset across different knowledge categories.
\paragraph{BioASQ}
The BioASQ~\citep{bioasq2023krithara} challenge, conducted every year, focuses on  techniques in large-scale biomedical semantic indexing and question answering (QA). For our experiments, we utilize Task B (Table \ref{tab:bioasq_samples}) from the eleventh edition of the BioASQ challenge (BioASQ 2023), which includes biomedical questions in English and their corresponding gold standard answers. We consider \textit{exact answers} as gold answers where available; otherwise, we refer to the \textit{ideal answers} field in the dataset.

\begin{table*}[ht]
    \centering
    \resizebox{\textwidth}{!}{ % Resize the table to fit the text width
    \begin{tabular}{|l|l|}
        \hline
        \textbf{Question}                                                           & \textbf{Answer}   \\ \hline
        Miami Beach in Florida borders which ocean?                                 & Atlantic          \\
        What was the occupation of Lovely Rita according to the song by the Beatles & Traffic Warden    \\
        Who was Poopdeck Pappys most famous son?                                    & Popeye            \\
        The Nazi regime was Germany's Third Reich; which was the first Reich?       & HOLY ROMAN EMPIRE \\ \hline
    \end{tabular}
    }
    \caption{Data samples from TriviaQA~\citep{joshi2017triviaqa}}
    \label{tab:triviaqa_samples}
\end{table*}

\begin{table*}[ht]
\centering
\resizebox{\textwidth}{!}{ % Resize the table to fit the text width
\begin{tabular}{|l|l|}
\hline
\textbf{Question}                                                                  & \textbf{Answer}                              \\ \hline
Which amino acid in implicated in the Blue diaper syndrome?                        & tryptophan                                   \\
What are the outcomes of ubiquitination?                                           & Protein degradation, Degradation of proteins \\
What causes Serpentine Supravenous Hyperpigmentation?                              & 5-fluorouracil, docetaxel                    \\
What are positive cell-cycle regulators that can cause cancer when mutated called? & Proto-oncogenes                              \\ \hline
\end{tabular}
}
\caption{Data samples from BioASQ~\citep{bioasq2023krithara}}
\label{tab:bioasq_samples}
\end{table*}

\subsection{Finetuning hyperparameters and Implementation}
\label{apdx:finetuning}

We fine-tune the models for generic causal language modeling (CAUSAL\_LM) task in autoregressive manner that predict the next token in a sequence based on the preceding tokens. In the CAUSAL\_LM task, labels are created directly from the prompt itself by using the subsequent tokens in the sequence as the target labels for prediction. For each position in the prompt sequence, the model takes the preceding tokens as input and the subsequent token as the label by progressively shifting the window of context during the fine-tuning process. We fine-tune our models for all experiments for 3 epochs using LoRA \citep{lora2022} with AdamW optmizer \citep{loshchilov2017decoupled}. We use an initial learning rate of $1e\hbox{-}4$, weight decay of $0.001$ and a warm up ratio of $0.03$. In our experiments we used Low-Rank Adaptation (LoRA) to efficiently fine-tune pre-trained LLMs and LVLMs for the causal language modeling task. For LLMs, we set the LoRA rank as 32, alpha parameter as 64 and a dropout of 0.1. LoRA was applied specifically to the following modules: \textit{q\_proj}, \textit{k\_proj}, \textit{v\_proj}, \textit{up\_proj}, and \textit{down\_proj}, resulting in less than 1\% of trainable model parameters (Table \ref{apdx_tab:trainable_parameters}). 
In addition to LoRA, we applied 4-bit normalized float (\textit{nf4}) quantization to the model's parameters and utilized \textit{FP16} precision during fine-tuning to reduce the computational overhead. 

For inference, we utilized \textit{FP16} precision and the default greedy decoding provided by Hugging Face with temperature value T=0.3. The predictive entropy and semantic entropy are estimated by generating 5 stochastic sequences from the model, each obtained through temperature sampling with a temperature setting of T=0.3. This temperature was chosen to obtain optimal uncertainty estimates balanced with high quality generated text, based on the ablation study shown in Figure \ref{apdx_fig:temperature_ablation}. Our source code was implemented using Pytorch\footnote{\href{https://pytorch.org/}{https://pytorch.org/}} framework and the models from Hugging Face\footnote{\href{https://huggingface.co/}{https://huggingface.co/}} library. For the UnLikelihood Training (ULT), we implemented the loss function as described in \citep{welleckneural} and fine-tuned it using LoRA, similar to the CLM and UA-CLM methods. For Calibration Tuning (CT) \citep{kapoor2024calibration} method, we utilized the calibration-tuned models provided in their repository\footnote{\href{https://github.com/activatedgeek/calibration-tuning}{https://github.com/activatedgeek/calibration-tuning}}. 

\begin{table}[t]
\centering
\resizebox{0.45\textwidth}{!}{%
\begin{tabular}{@{}lll@{}}
\toprule
Model &  & Percentage of trainable parameters \\ \midrule
Llama-2-7B &  & 0.83\% \\
Llama-2-13B &  & 0.67\% \\
Gemma-2B &  & 1.03\% \\ \bottomrule
\end{tabular}%
}
\caption{Percentage of trainable parameters when fine-tuning with LoRA in our experiments}
\label{apdx_tab:trainable_parameters}
\end{table}

For the LVLM model, LLaVA-1.5~\citep{llava2024liu}, we configured LoRA with a rank of 8, an alpha value of 8, and applied a 0.1 dropout rate to mitigate overfitting on the small OK-VQA training subset. In addition to the proposed UA-CLM loss, we experimented with a combined loss function that anneals the CLM loss with our UA-CLM loss. This approach allows the model to learn to answer OK-VQA queries using the context provided in the early stages of training, without uncertainty calibration. As training progresses, we shift our focus toward calibrating the model's uncertainty. By this stage, the model has already learned to answer visual question-answering prompts, allowing us to refine its performance on questions it is likely to answer correctly or incorrectly, based on insights gained during the initial training phases. Specifically, we assign a higher weight to the CLM loss in the early stages of training, gradually increasing the weight of the UA-CLM loss after 20\% of the training is completed as shown in Equation \ref{eq:okvqa_annealed_loss}. Our ablation results for this study are presented in Table \ref{apdx_tab:ablation_loss}.
\begin{equation}
\mathcal{L} = \mathcal{L_{\mathrm{CLM}}} + \beta \cdot \mathcal{L_{\mathrm{UA\hbox{-}CLM}}} 
\label{eq:okvqa_annealed_loss}
\end{equation}
\text{where } 
\begin{equation}
\beta = 
\begin{cases} 
0.2 & \text{if } \text{steps} \leq 0.2 \cdot \text{total\_steps} \\
0.8 & \text{if } \text{steps} > 0.2 \cdot \text{total\_steps} 
\end{cases}
\end{equation}

\subsection{Prompt Template\label{apdx:prompt}}

Open-book QA Prompt (CoQA):
\begin{quote}

Answer the following question as briefly as possible. 

Context:  [Provided context paragraph] \\
Question: [Associated Question] \\
Answer: 
\end{quote}

Closed-book QA Prompt (TriviaQA and BioASQ):
\begin{quote}

Answer the following question as briefly as possible. 

Question: [Question] \\
Answer: 
\end{quote}

Biography generation Prompt (BioGen):
\begin{quote}
You are an AI assistant. You use a tone that is technical and scientific. \\
USER: Write a paragraph for [name]'s biography. \\
ASSISTANT:
\end{quote}

\begin{table*}[H]
\centering
\resizebox{0.7\textwidth}{!}{%
\begin{tabular}{@{}lllll@{}}
\toprule
Dataset &  & Text generation lenth &  & eos\_tokens \\ \midrule
TriviaQA and OKVQA &  & phrase-length generation &  & {[}'\textbf{.}',   '\textbf{,}', '\textbf{\textbackslash{}n}', tokenizer.eos\_token\_id{]} \\
CoQA and BioASQ &  & sentence-length generation &  & {[}'\textbf{.}',   '\textbf{\textbackslash{}n}', tokenizer.eos\_token\_id{]} \\
BioGen &  & paragraph-length generation &  & {[}'\textbf{\textbackslash{}n}',   tokenizer.eos\_token\_id{]} \\ \bottomrule
\end{tabular}%
}
\caption{End-of-Sequence tokens ($eos\_token$) settings for text generation }
\label{apdx_tab:eos_tokens}
\end{table*}

\section{Fine-tuning Latency comparison analysis}
 The training latency and throughput comparison is presented in Table \ref{apdx_tab:latency}. The proposed UA-CLM method takes similar fine-tuning time as the standard CLM method, while providing the benefits of improved uncertainty calibration. The latency measurements for both methods were conducted on similar setup for a fair comparison: a batch size of 2 with an average of 512 tokens per sequence, 4-bit normalized float (nf4) quantization for model parameters, and FP16 precision for computation on Intel(R) Xeon(R) Gold 6448H CPU and Nvidia A100 80G GPU, with LoRA fine-tuning (less than 1\% model parameters updated), as described in Section \ref{apdx:finetuning}. While the numbers may vary with different setups, the relative comparison should ideally remain consistent.

\begin{table*}[ht]
\centering
\resizebox{0.75\textwidth}{!}{%
\begin{tabular}{@{}lllllllll@{}}
\toprule
Model &  & Method &  & Latency &  & Throughput &  & Relative Latency \\ \midrule
\multirow{2}{*}{Llama-2-7B} &  & CLM &  & 2.63 ms/token &  & 379 tokens/s &  & 1x \\
 &  & UA-CLM &  & 2.78 ms/token &  & 359 tokens/s &  & 1.06x \\
 &  &  &  &  &  &  &  & \\ \midrule \\
\multirow{2}{*}{Llama-2-13B} &  & CLM &  & 4.50 ms/token &  & 222 tokens/s &  & 1x \\
 &  & UA-CLM &  & 4.85 ms/token &  & 206 tokens/s &  & 1.07x \\ \bottomrule 
\end{tabular}%
}
\caption{Latency comparison between CLM and UA-CLM fine-tuning methods}
\label{apdx_tab:latency}
\end{table*}

\section{Text generation quality metrics}
\label{apdx:text_eval_metrics}
\begin{itemize}
    \item \textbf{ROUGE-L}~\citep{lin2004automatic}: Recall-Oriented Understudy for Gisting Evaluation (ROUGE) is a widely-used evaluation metric for assessing the quality of text generated based on n-gram matching. We use the Rouge-L variant which uses the longest common subsequence between the generated answer and the ground truth answer. 
    \item \textbf{Exact Match (EM): } Exact Match (EM) metric is a stringent evaluation criterion used to assess the performance of models on tasks such as question answering (QA), where a generated response is compared to a reference answer. It is a widely used metric for open-book QA,  this metric evaluates a model's ability to extract the precise text span from the context to answer a question.
    \item \textbf{Accuracy: } The generated answer is considered as accurate if it achieves Rouge-L$(y, \hat{y}) > 0.3$, for a given reference answer $y$ and a model generation $\hat{y}$. We follow this criterion for quantifying accuracy in free-form text generation based on the findings from ~\citep{kuhn2023semantic} that demonstrated this criterion closely matches the human evaluation accuracy on COQA and TriviaQA datasets, both of which are utilized in our experiments.
    \item \textbf{BERTScore} \citep{zhang2019bertscore}:  BERTScore utilizes word embeddings to compute a similarity score between the tokens in the prediction and ground truth and has shown to well correlate with human judgement. We report Precision, Recall and F1 BERTScores for all our experiments.
    %\item \textbf{METEOR \citep{banerjee2005meteor}: } A commonly used metric in machine translation evaluation. METEOR is based on unigram matching and also takes into account the ordering of words in the generation and the reference. 
\end{itemize}

\section{Uncertainty estimation metrics}
We assess uncertainty in natural language predictions by utilizing the Area Under the Receiver Operating Characteristic (AUROC) scores, calculated between correct and incorrect predictions across the following metrics:
\begin{itemize}
    %\item \textbf{Max Token Probability}: Probability of the predicted token is a direct indicator of the model's confidence. 
    \item \textbf{Predictive Entropy}~\cite{fomicheva2020unsupervised}: This is a widely used measure for uncertainty estimation and is defined as the entropy of the model's output probability distribution from stochastic generated responses.
    Formally, for a specific instance $x$, the predictive entropy, denoted as $P_{E}(x)$, is defined as the conditional entropy of the output random variable $Y$, with realization $y$, given $x$ \citep{kuhn2023semantic}: $P_{E}(x) = H(Y|x) = -\int p(y|x) \ln p(y|x) dy$
    
    \item \textbf{Perplexity}~\cite{fomicheva2020unsupervised}: A standard metric to assess the quality of model and is defined as the inverse probability of the generated text: 
    $ \text{Perplexity} = \exp\left(-\frac{1}{N} \sum_{i=1}^N \log_2 p(w_i|w_1, \ldots, w_{i-1})\right) $

    \item \textbf{Semantic Entropy} \citep{kuhn2023semantic}: Defined as entropy of output distributions in semantic event-space rather than traditional token event-space and has been shown to be a good indicator in detecting confabulation in language models.
\end{itemize}

\section{Additional Results}
\label{apdx:additional_results}

\subsection{Effect of temperature on quality of generated text and uncertainty quantification metrics: An ablation study}
\label{apdx:ablation_temperature}
\begin{figure*}[ht]
\centering
\includegraphics[width=1.0\textwidth]{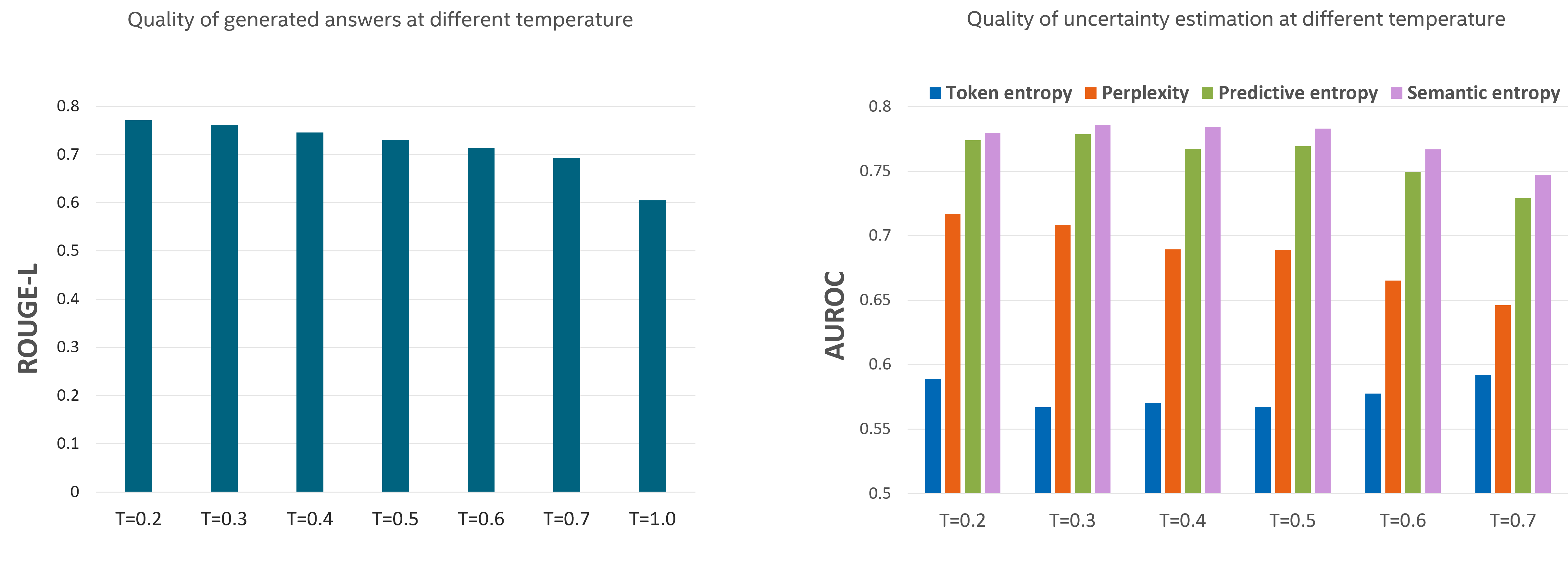}
\caption{\small Ablation study: Effect of temperature value on the quality of generated text and the quality of uncertainty estimates evaluated with AUROC for hallucination detection. The study was performed on pre-trained Llama-2-7B model with CoQA dataset. Based on this study, we selected temperature T=0.3 as it results in optimal AUROC and ROUGE-L scores.}
\label{apdx_fig:temperature_ablation}
\end{figure*}

The ablation study conducted on the pre-trained Llama-2-7B model using the CoQA dataset provides insightful findings regarding the impact of temperature values on text generation quality and uncertainty estimation as shown in Figure \ref{apdx_fig:temperature_ablation}. By systematically varying the temperature parameter, the study aimed to identify the optimal setting that balances the quality of generated text with effective hallucination detection, as measured by AUROC. The results revealed that a temperature value of T=0.3 achieved the best performance, yielding optimal AUROC scores for uncertainty estimates and ROUGE-L scores for text quality. 

%\begin{figure}
%    \centering
    %\includegraphics[width=13cm]{./figures/legend_line.png}
%    \includegraphics[width=6.5cm]{./figures/selective_gen_llama-2-7b_triviaqa_rougeL.png}
%    \qquad
%    \includegraphics[width=6.5cm]{./figures/selective_gen_llama-2-7b_triviaqa_accuracy.png}
%    \caption{Selective generation (Llama-2-7B/TriviaQA)}%
%    \label{apdx_fig:selective_gen}
%\end{figure}

\subsection{Evaluation of various text generation quality metrics}
\label{apdx:text_generation_quality_metrics}

The results in the Table~\ref{app_tab:text_gen_quality_eval} presents a detailed quantitative evaluation of various text generation quality metrics across various models and datasets. It compares standard Causal Language Modeling (CLM) with our Uncertainty-Aware Causal Language Modeling (UA-CLM).

\begin{table*}[ht]
\centering
\caption{\small Evaluation of generated text quality metrics: Comparative analysis of Causal Language Modeling (CLM) and Uncertainty-aware Causal Language Modeling (UA-CLM) fine-tuning methods. The results in the table indicate that UA-CLM achievies similar or better generated text quality metrics than standard CLM across a range of models and datasets.}
\resizebox{0.95\textwidth}{!}{%
\begin{tabular}{@{}llllcccccc@{}}
\toprule
Dataset & Model & \begin{tabular}[c]{@{}l@{}}Finetuning \\ Method\end{tabular} &  & Rouge-L & \begin{tabular}[c]{@{}c@{}}Exact Match \end{tabular} & \begin{tabular}[c]{@{}c@{}}Accuracy \end{tabular} & \begin{tabular}[c]{@{}c@{}}BERT Score \\ (Precision)\end{tabular} & \begin{tabular}[c]{@{}c@{}}BERT Score \\ (Recall)\end{tabular} & \begin{tabular}[c]{@{}c@{}}BERT Score \\ (F1)\end{tabular} \\ \midrule
% &  &  &  &  &  &  &  &  &  \\
\multirow{8}{*}{CoQA} & \multirow{2}{*}{Llama-2-7b} & CLM &  & 0.8886 & 0.8071 & 0.9253 & 0.9633 & 0.9598 & 0.9604 \\ [0.05cm]
 &  & UA-CLM &  & 0.8882 & 0.8027 & 0.9264 & 0.9671 & 0.9644 & 0.9648 \\ 
 \cmidrule(l){2-10} 
% &  &  &  &  &  &  &  &  &  \\
 & \multirow{2}{*}{Llama-2-13b} & CLM &  & 0.9106 & 0.8434 & 0.9406 & 0.9678 & 0.9639 & 0.9650 \\ [0.05cm]
 &  & UA-CLM &  & 0.9118 & 0.8204 & 0.9461 & 0.9732 & 0.9698 & 0.9705 \\
 \cmidrule(l){2-10} 
 %&  &  &  &  &  &  &  &  &  \\
 & \multirow{2}{*}{Gemma-2b} & CLM &  & 0.8654 & 0.7606 & 0.9143 & 0.962 & 0.9548 & 0.9570 \\ [0.05cm]
 &  & UA-CLM &  & 0.8632 & 0.7632 & 0.9088 & 0.9627 & 0.9554 & 0.9578 \\
 \midrule
 %&  &  &  & \multicolumn{1}{l}{} & \multicolumn{1}{l}{} & \multicolumn{1}{l}{} & \multicolumn{1}{l}{} & \multicolumn{1}{l}{} & \multicolumn{1}{l}{} \\
\multirow{8}{*}{TriviaQA} & \multirow{2}{*}{Llama-2-7b} & CLM &  & 0.5867 & 0.4939 & 0.6385 & 0.8743 & 0.8785 & 0.8754 \\ [0.05cm]
 &  & UA-CLM &  & 0.6342 & 0.5627 & 0.6754 & 0.8951 & 0.8883 & 0.8910 \\
 \cmidrule(l){2-10} 
% &  &  &  &  &  &  &  &  &  \\
 & \multirow{2}{*}{Llama-2-13b} & CLM &  & 0.6588 & 0.5883 & 0.6967 & 0.9026 & 0.8989 & 0.9001 \\ [0.05cm]
 &  & UA-CLM &  & 0.7277 & 0.6445 & 0.7710 & 0.9204 & 0.9164 & 0.9177 \\
 \cmidrule(l){2-10} 
% &  &  &  &  &  &  &  &  &  \\
 & \multirow{2}{*}{Gemma-2b} & CLM &  & 0.4349 & 0.3674 & 0.4759 & 0.8375 & 0.8349 & 0.8355 \\ [0.05cm]
 &  & UA-CLM &  & 0.4563 & 0.3915 & 0.4959 & 0.8404 & 0.8382 & 0.8387 \\
 \midrule
% &  &  &  &  &  &  &  &  &  \\
\multirow{2}{*}{OK-VQA} & \multirow{2}{*}{Llava-1.5-7b} & CLM &  & 0.5569 & 0.5099 & 0.5891 & 0.8897 & 0.8864 & 0.8877 \\ [0.05cm]
 &  & UA-CLM &  & 0.5354 & 0.4950 & 0.5643 & 0.8841 & 0.8820 & 0.8827 \\ \bottomrule
\end{tabular}%
}
\label{app_tab:text_gen_quality_eval}
\end{table*}

\begin{table*}[ht]
\centering
\caption{\small Uncertainty calibration analysis: The results show UA-CLM have more pronounced negative correlation between the uncertainty estimates and the generated text quality (ROUGE-L) than standard Causal Language Modeling CLM, indicating enhanced reliability in uncertainty quantification with UA-CLM.}
\resizebox{0.95\textwidth}{!}{%
\begin{tabular}{@{}lllcccccccccc@{}}
\toprule
\multirow{2}{*}{Dataset} & \multirow{2}{*}{Model} & \multirow{2}{*}{\begin{tabular}[c]{@{}l@{}}Finetuning \\ Method\end{tabular}} & \multicolumn{4}{c}{Spearman's rank correlation coefficient $\downarrow$} & \multicolumn{1}{l}{} & \multicolumn{1}{l}{} & \multicolumn{4}{c}{Pearson correlation coefficient $\downarrow$} \\ \cmidrule(l){4-7} \cmidrule(l){9-13} 
 &  &  & \begin{tabular}[c]{@{}c@{}}Token \\ Entropy\end{tabular} & Perplexity & \begin{tabular}[c]{@{}c@{}}Predictive \\ Entropy\end{tabular} & \begin{tabular}[c]{@{}c@{}}Semantic \\ Entropy\end{tabular} &  &  & \begin{tabular}[c]{@{}c@{}}Token \\ Entropy\end{tabular} & Perplexity & \begin{tabular}[c]{@{}c@{}}Predictive \\ Entropy\end{tabular} & \begin{tabular}[c]{@{}c@{}}Semantic \\ Entropy\end{tabular} \\ \midrule
% &  &  &  &  &  &  &  &  &  &  &  &  \\
\multirow{8}{*}{CoQA} & \multirow{2}{*}{Llama-2-7b} & CLM & -0.2130 & -0.2379 & -0.3398 & -0.2898 &  &  & -0.2029 & -0.2109 & -0.2710 & -0.2881 \\ [0.05cm]
 &  & UA-CLM & \textbf{-0.2479} & \textbf{-0.3401} & \textbf{-0.4334} & \textbf{-0.3742} &  &  & \textbf{-0.3414} & \textbf{-0.3414} & \textbf{-0.3414} & \textbf{-0.3414} \\
 \cmidrule(r){3-13}
% &  &  &  &  &  &  &  &  &  &  &  &  \\
 & \multirow{2}{*}{Llama-2-13b} & CLM & -0.2325 & -0.2523 & -0.3253 & -0.3004 &  &  & -0.2302 & -0.2495 & -0.3001 & -0.2636 \\ [0.1cm]
 &  & UA-CLM & \textbf{-0.2398} & \textbf{-0.3280} & \textbf{-0.4170} & \textbf{-0.3717} & \textbf{} & \textbf{} & \textbf{-0.2335} & \textbf{-0.3244} & \textbf{-0.3269} & \textbf{-0.3481} \\
  \cmidrule(r){3-13}
% &  &  &  &  &  &  &  &  &  &  &  &  \\
 & \multirow{2}{*}{Gemma-2b} & CLM & -0.3639 & -0.3629 & -0.4335 & -0.3756 &  &  & -0.3860 & -0.3713 & -0.3483 & -0.3399 \\ [0.05cm]
 &  & UA-CLM & \textbf{-0.3676} & \textbf{-0.4063} & \textbf{-0.4476} & \textbf{-0.4127} &  &  & \textbf{-0.4033} & \textbf{-0.4019} & \textbf{-0.3517} & \textbf{-0.3530} \\
 \midrule
% &  &  & \multicolumn{1}{l}{} & \multicolumn{1}{l}{} & \multicolumn{1}{l}{} & \multicolumn{1}{l}{} & \multicolumn{1}{l}{} & \multicolumn{1}{l}{} & \multicolumn{1}{l}{} & \multicolumn{1}{l}{} & \multicolumn{1}{l}{} & \multicolumn{1}{l}{} \\
\multirow{8}{*}{TriviaQA} & \multirow{2}{*}{Llama-2-7b} & CLM & -0.5627 & -0.5863 & -0.5765 & -0.5994 &  &  & -0.5047 & -0.4854 & -0.2864 & -0.5020 \\ [0.05cm]
 &  & UA-CLM & \textbf{-0.5713} & \textbf{-0.6011} & \textbf{-0.5822} & -0.5980 &  &  & \textbf{-0.5385} & \textbf{-0.5326} & \textbf{-0.3382} & -0.4916 \\
  \cmidrule(r){3-13}
% &  &  &  &  &  &  &  &  &  &  &  &  \\
 & \multirow{2}{*}{Llama-2-13b} & CLM & -0.5711 & -0.5845 & -0.5522 & -0.5959 &  &  & -0.5155 & -0.4915 & -0.4548 & -0.4612 \\ [0.1cm]
 &  & UA-CLM & \textbf{-0.5725} & \textbf{-0.5862} & \textbf{-0.5607} & -0.5854 &  &  & \textbf{-0.5362} & \textbf{-0.5407} & \textbf{-0.4786} & -0.4479 \\
  \cmidrule(r){3-13}
% &  &  &  &  &  &  &  &  &  &  &  &  \\
 & \multirow{2}{*}{Gemma-2b} & CLM & -0.5636 & -0.5772 & -0.5609 & -0.5537 &  &  & -0.5020 & -0.4534 & -0.4494 & -0.4514 \\ [0.1cm]
 &  & UA-CLM & -0.5623 & \textbf{-0.5913} & -0.5457 & \textbf{-0.5928} &  &  & \textbf{-0.5164} & \textbf{-0.5010} & \textbf{-0.4534} & \textbf{-0.4947} \\
 \midrule
% &  &  &  &  &  &  &  &  &  &  &  &  \\
\multirow{2}{*}{OK-VQA} & \multirow{2}{*}{Llava-1.5-7b} & CLM & -0.1253 & -0.1132 & -0.1320 & -0.1062 &  &  & -0.0862 & -0.0861 & -0.1256 & -0.1340 \\ [0.05cm]
 &  & UA-CLM & \textbf{-0.1606} & \textbf{-0.1619} & \textbf{-0.2050} & \textbf{-0.2660} &  &  & -0.0748 & \textbf{-0.1214} & \textbf{-0.2100} & \textbf{-0.3020} \\ \bottomrule
\end{tabular}%
}
\label{app_tab:correlation_eval}
\end{table*}

\subsection{Correlation between uncertainty quantification and quality of generated text}
\label{apdx:correlation_metrics}
\begin{figure*}[htbp]
\centering
\includegraphics[width=0.9\textwidth]{./figures/legend.png}
\includegraphics[width=1.0\textwidth]{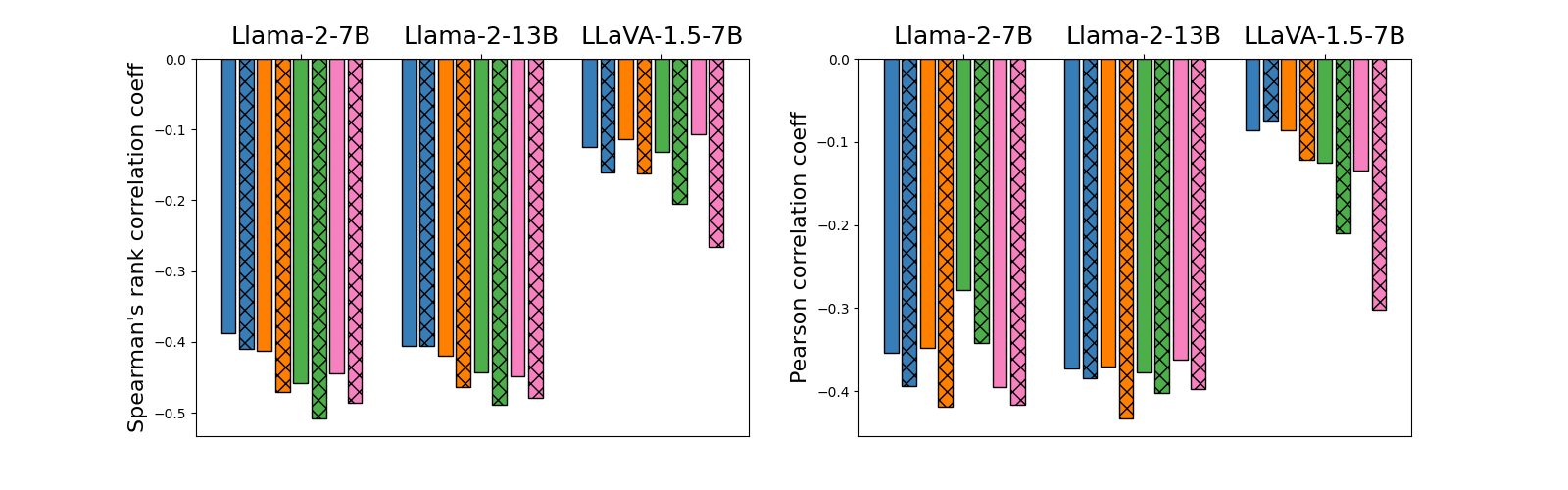}
\caption{Uncertainty calibration analysis: Spearman's rank correlation coefficient and Pearson correlation coefficient between uncertainty estimates and generated text quality (ROUGE-L) scores for free-form open-ended question answering. Stronger negative correlation is desired for well-calibrated uncertainty quantification.}
\label{fig:correlation_bar_plot}
\end{figure*}

Figure \ref{fig:correlation_bar_plot} and Table~\ref{app_tab:correlation_eval} presents quantitative data with the values of Spearman's rank correlation coefficient~\citep{zwillinger1999crc} and Pearson correlation coefficient~\citep{benesty2009pearson} across different models, datasets, and uncertainty quantification (UQ) metrics, with a specific focus on comparing standard Causal Language Modeling (CLM) and our Uncertainty-Aware Causal Language Modeling (UA-CLM). The data reveals that UA-CLM exhibits a stronger inverse correlation between UQ metrics and ROUGE-L scores, indicating better reliability of uncertainty estimates. This enhanced inverse relationship suggests that UA-CLM is more adept at associating higher uncertainty with low quality text generation quality and vice versa, which is a key indicator of better uncertainty calibration. 

\begin{table*}[htbp]
\centering
\caption{Ablation study: Effect of different loss functions and annealing during fine-tuning. Exact match is used as accuracy metric in computing AUARC.}
\resizebox{0.9\textwidth}{!}{%
\begin{tabular}{@{}lllcccccccccc@{}}
\toprule
\multirow{2}{*}{Dataset} & \multirow{2}{*}{Model} & \multirow{2}{*}{Fine-tuning Loss} & \multicolumn{4}{c}{AUROC   (Hallucination/Confabulation detection)} &  &  & \multicolumn{4}{c}{AUARC (Area under   rejection accuracy curve)} \\ \cmidrule(l){4-7} \cmidrule(l){9-13} 
 &  &  & \begin{tabular}[c]{@{}c@{}}Token \\ Entropy\end{tabular} & Perplexity & \begin{tabular}[c]{@{}c@{}}Predictive \\ Entropy\end{tabular} & \begin{tabular}[c]{@{}c@{}}Semantic \\ Entropy\end{tabular} &  &  & \begin{tabular}[c]{@{}c@{}}Token \\ Entropy\end{tabular} & Perplexity & \begin{tabular}[c]{@{}c@{}}Predictive \\ Entropy\end{tabular} & \begin{tabular}[c]{@{}c@{}}Semantic \\ Entropy\end{tabular} \\ \midrule
 &  &  &  &  &  &  &  &  &  &  &  &  \\
\multirow{3}{*}{OKVQA} & \multirow{3}{*}{Llava-1.5-7b} & $\mathcal{L_{\mathrm{CLM}}}$ & 0.5504 & 0.5419 & 0.5455 & 0.5370 &  &  & 0.5809 & 0.5781 & 0.5790 & 0.5747 \\
 &  & $\mathcal{L_{\mathrm{UA\hbox{-}CLM}}}$ & 0.5839 & 0.6032 & 0.5701 & 0.6727 &  &  & 0.5657 & 0.5771 & 0.5601 & 0.6028 \\
 &  & $\mathcal{L_{\mathrm{CLM}}} + \beta * \mathcal{L_{\mathrm{UA\hbox{-}CLM}}}$ & 0.6001 & 0.5984 & 0.6106 & 0.6638 &  &  & 0.5989 & 0.5965 & 0.6012 & 0.6265 \\ \midrule
 &  &  &  &  &  &  &  &  &  &  &  &  \\
\multirow{2}{*}{CoQA} & \multirow{2}{*}{Llama-2-7b} & $\mathcal{L_{\mathrm{CLM}}}$ & 0.6252 & 0.632 & 0.6635 & 0.6889 &  &  & 0.8230 & 0.8290 & 0.8516 & 0.8405 \\
 &  & $\mathcal{L_{\mathrm{UA\hbox{-}CLM}}}$ & 0.6955 & 0.7398 & 0.7413 & 0.7741 &  &  & 0.8246 & 0.8477 & 0.8743 & 0.8571 \\
 &  & $\mathcal{L_{\mathrm{CLM}}} + \beta * \mathcal{L_{\mathrm{UA\hbox{-}CLM}}}$ & 0.6101 & 0.6183 & 0.6978 & 0.7252 &  &  & 0.8153 & 0.8153 & 0.8614 & 0.8455 \\\midrule
 &  &  &  &  &  &  &  &  &  &  &  &  \\
\multirow{3}{*}{TriviaQA} & \multirow{3}{*}{Llama-2-13b} & $\mathcal{L_{\mathrm{CLM}}}$ & 0.8264 & 0.8333 & 0.7971 & 0.8407 &  &  & 0.7464 & 0.7526 & 0.7532 & 0.7556 \\
 &  & $\mathcal{L_{\mathrm{UA\hbox{-}CLM}}}$ & 0.8297 & 0.8352 & 0.8033 & 0.8447 &  &  & 0.7960 & 0.8059 & 0.804 & 0.8069 \\
 &  & $\mathcal{L_{\mathrm{CLM}}} + \beta * \mathcal{L_{\mathrm{UA\hbox{-}CLM}}}$ & 0.8340 & 0.8263 & 0.8049 & 0.8307 &  &  & 0.7666 & 0.7692 & 0.7673 & 0.7693 \\ \bottomrule
\end{tabular}%
}
\\
\label{apdx_tab:ablation_loss}
\end{table*}

%Figure~\ref{apdx_fig:selective_gen} shows results on selective generation, based on varying levels of abstaining from providing generated response informed by uncertainty estimates. We plotted both ROUGE-L scores and accuracy as functions of the abstention rate, showing how the models perform as they increasingly withhold responses in situations of high uncertainty. The plots clearly shows that the UA-CLM outperforms CLM across all the four uncertainty metrics.

%\subsection{Choice of BioASQ for Out-of-domain prompt detection task}
%\label{apdx:out_of_domain_detection}
%We selected the BioASQ dataset for the out-of-domain detection task after validating that the text generation quality metrics on a pre-trained Llama-2 model were extremely low compared to the other datasets used in our study (CoQA, TriviaQA), as shown in Table \ref{apdx_tab:bioasq_text_quality}. It is not possible to definitively determine if BioASQ data was part of the Llama-2 model's pre-training data, as this information is not publicly available. However, the validation from the results in table below suggested that the BioASQ data could possibly be out of the Llama-2 model's pre-training knowledge.

\end{document}